\documentclass[lettersize,journal]{IEEEtran}
\usepackage{verse}
\usepackage{amsmath,amsfonts}
\usepackage{algorithmic}
\usepackage{algorithm}
\usepackage{array}
\usepackage[caption=false,font=normalsize,labelfont=sf,textfont=sf]{subfig}
\usepackage{textcomp}
\usepackage{stfloats}
\usepackage{url}
\usepackage{verbatim}
\usepackage{graphicx}
\usepackage{cite}
\IEEEoverridecommandlockouts
\usepackage{cite}
\usepackage{amsmath,amssymb,amsfonts}
\usepackage{algorithmic}
\usepackage{graphicx}
\usepackage{textcomp}
\usepackage{xcolor}
\usepackage{amsbsy}
\usepackage{stmaryrd}
\usepackage{bm}
\usepackage{algorithm}
\usepackage{booktabs}
\usepackage{caption}
\usepackage{stfloats}
\usepackage{multirow}
\usepackage{tikz}

\def\BibTeX{{\rm B\kern-.05em{\sc i\kern-.025em b}\kern-.08em
    T\kern-.1667em\lower.7ex\hbox{E}\kern-.125emX}}

\begin{document}

\title{Online Multi-Label Classification under Noisy and Changing Label Distribution}

\author{YizhangZou, XuegangHu,~\IEEEmembership{Staff,~IEEE}, PeipeiLi,~\IEEEmembership{Staff,~IEEE}, JunHu, YouWu

\thanks{
    YizhangZou, YouWu, XuegangHu, Peipei Li are with the Key Laboratory of Knowledge Engineering with Big Data, Ministry of Education, School of Computer Science and Information Engineering, Hefei University of Technology, Hefei, China. (E-mail: \{yizhangzou, wuyou\}@mail.hfut.edu.cn,
    \{jsjxhuxg, peipeili\}@hfut.edu.cn)
}

\thanks{
    JunHu is with the School of Computing, National University of Singapore, Singapore. (E-mail: jun.hu@nus.edu.sg)
}

}



\maketitle

\begin{abstract}
Multi-label data stream usually contains noisy labels in the real-world applications, namely occuring in both relevant and irrelevant labels. However, existing online multi-label classification methods are mostly limited in terms of label quality and fail to deal with the case of noisy labels. On the other hand, the ground-truth label distribution may vary with the time changing, which is hidden in the observed noisy label distribution and difficult to track, posing a major challenge for concept drift adaptation. Motivated by this, we propose an online multi-label classification algorithm under Noisy and Changing Label Distribution (NCLD).  The convex objective is designed to simultaneously model the label scoring and the label ranking for high accuracy, whose robustness to NCLD benefits from three novel works: 1) The local feature graph is used to reconstruct the label scores jointly with the observed labels, and an unbiased ranking loss is derived and applied to learn reliable ranking information. 2) By detecting the difference between two adjacent chunks with the unbiased label cardinality, we identify the change in the ground-truth label distribution and reset the ranking or all information learned from the past to match the new distribution. 3) Efficient and accurate updating is achieved based on the updating rule derived from the closed-form optimal model solution. Finally, empirical experimental results validate the effectiveness of our method in classifying instances under NCLD.
\end{abstract}


\section{Introduction}
\IEEEPARstart{O}nline multi-label classification (OMC) aims to instantly annotate multi-label streaming objects that arrive sequentially, which are associated with two or more labels \cite{FirstOrder,WeightedEnsemble}. Multi-label data stream is very common in real-world applications. For example, the online image recognition system automatically annotates multiple objects in the streaming images \cite{MLDN}; in movie recommendation, the online customers accessing the movie website are recommended several movies of different types, e.g., tragedy, science fiction and horror movies \cite{OMC}; in music emotion classification, different types of emotions are recognized from continuous pieces of music \cite{emotions}.

Pure and noiseless multi-label data is rarely available in the online environment due to labelling costs and human error \cite{ICMR,Est}, noisy labels are often present in the label set and appear in relevant or irrelevant labels, leading to performance degradation. In particular, existing OMC works aim to instantly obtain accurate continuous label outputs, and partition the outputs to provide reliable predictions over a reasonable threshold. For example, the work of \cite{RELM} integrates the label ranking information and the label scoring information into an online learning framework, where the former provides an accurate estimated ranking between labels and the latter ensures the rationality of using zero as a fixed threshold. However, in the case of noisy labels, incorrect labels and corresponding ranking information will mislead the model, and the error will propagate over time, causing inaccurate and unstable OMC performance. Though some new OMC works have been proposed recently to tackle the limited-supervision case such as semi-supervised classification \cite{semi1} and classification with missing labels \cite{Balance}, the OMC with noisy labels is first formulated and addressed to the extent of our knowledge.

Furthermore, as a widespread type of concept drift, the changes in the distribution of labels are formulated and addressed by previous OMC works \cite{Imba,CD1}. Together with the presence of noisy labels, real-world applications raise the need to perform online classification under noisy and changing label distribution (NCLD). In line with \cite{RELM}, throughout the paper we also consider the two extreme cases of concept drift in the ground-truth label distribution: 1) from a single-label distribution to a multi-label distribution (i.e., concept growth); 2) from a multi-label distribution to a single-label distribution (i.e., concept reduction). Note that under NCLD, only the noisy label distribution is observed and available instead of the real distribution, resulting in the performance degradation of previous methods that require the clean labels for the detection and adaptation to distribution changes. For example of concept growth, when the single-label distribution drifts to the multi-label distribution, existing accuracy-based detection methods \cite{Reacting,AptE} can not give an indication of drift because of the incalculable classification loss utilizing the unavailable ground-truth labels. Similarly, distribution-based detection methods \cite{En3,Prototype} also fail because the label cardinality increase \cite{Review} brought by concept growth cannot be measured based on the noisy label distribution, and it is also difficult to track and adapt to the ground-truth label distribution for the adaptive OMC methods without concept drift detection \cite{RELM,semi1}.

In conclusion, there is a gap in considering streaming noisy labels for OMC, let alone performing accurate OMC under noisy and changing label distribution (NCLD). Motivated by this, we propose a novel NCLD-oriented OMC method. As stated in \cite{FALT,MOPAL}, we also formulate the convex objective which incorporates both label scoring and ranking regularization component for the verified high performance. Since the noise-induced label scoring and ranking information will mislead the model as mentioned above, it is necessary to enhance the robustness of label scoring and ranking regularization terms to noisy labels respectively. Additionally, detecting and adapting to the ground-truth concept drift also matters in classification under NCLD. To reduce the time complexity of our method for high-dimensional data, the Extreme learning machine (ELM) framework is used due to its high efficiency. ELM is a single-layer feedforward neural network that reduces the high dimensionality via a non-linear mapping with the random input weights, which is widely used in online learning \cite{ELM,semi2}. In this paper, a novel ELM-based online multi-label classification method under NCLD is proposed, which utilizes the label scoring and ranking regularized framework. The contributions of this paper are summarized as follows:

\begin{itemize}
    \item Based on the idea of label reconstruction, we develop the new label scoring term, which scores the label outputs of each instance jointly taking into account the observed labels and the label scores of its local nearest neighbours, ensuring that the scores of noiseless labels are more credible than noisy labels for better model fitting. 
  
    \item Using the unbiased estimator, we derive an unbiased label ranking term with respect to the ranking loss under the noiseless case, thus constructing the robust ranking order between the ground-truth relevant and irrelevant labels.
      
    \item Under NCLD, we derive an unbiased statistic with respect to label cardinality \cite{Review} to detect ground-truth concept drift, and propose two adaptation strategies to maintain robust performance during concept drift.

    \item We derive an efficient closed-form solution for the unified objective and the corresponding sequential update rule, thus preserving all information learned from the past without storing it for the online model. 
\end{itemize}

With contributions 1) and 2), our work achieves that the scores of real relevant labels and irrelevant labels can follow the bipartition over the fixed threshold of 0. Thus, during the online classification process, we can choose 0 as a fixed threshold to obtain robust binary predictions under noisy label distributions. Contributions 1), 2) and 3) together provide robustness to both noisy labels and concept drift (i.e., NCLD). Finally, contribution 4) enables model efficiency and accuracy, where efficiency is guaranteed by the ELM framework and the closed-form solution ensures competitive accuracy compared to batch versions.

The rest of this paper is organized as follows: Section \uppercase\expandafter{\romannumeral 2} reviews related work on OMC and online classification with limited supervision. Section \uppercase\expandafter{\romannumeral 3} gives a detailed introduction to our method. Next, Section \uppercase\expandafter{\romannumeral 4} presents the experimental results and Section \uppercase\expandafter{\romannumeral 5} concludes the paper by summarizing our findings and suggesting possible directions for future research. Finally, Section \uppercase\expandafter{\romannumeral 6} gives the detailed proofs of the propositions proposed in the paper.

\section{Related Work}
The related arts of our work are about online multi-label classification and online classification with limited supervision \cite{Trends}, we briefly review the representative works in these two areas.

\subsection{Online Multi-Label Classification}
Online multi-label classification performs instant annotation on multi-label objects arriving in sequence, which can be broadly categorized into three types, namely replay-based methods, regularization-based methods and ensemble methods. As a lazy learning-based method, replay-based methods maintain a data summary that approximates the latest data distribution and train the model based on it when a new instance comes, the problem of concept drift can be properly tackled since the data summary is dynamically updated. The work of \cite{MCIC} acts online classification by utilizing a weighted clustering model, which considers the change of data distribution. OnSeMl \cite{semi1} constructs two buffer pools storing the latest instances with or without label annotations respectively to perform semi-supervised online classification.  Regularization-based methods continuely update the online model based on the model induced by the past data and the current incoming data. The work of \cite{Binary} transforms the labels into a continuous value and applies a regression algorithms to perform online classification. Aiming at detecting the potential emerging labels, MuENL \cite{Emerge} sequentially updates the 
old classifier built for known labels and collaboratively builds the new classifier for each new label. Ensemble methods, different from single-model methods, maintain and update multiple models with high classification accuracy on classifying the latest data, and integrate the multiple model outputs as the final classification results. The work of \cite{NovelEnsemble} introduced 
dynamically-weighted stacked ensemble learning method to assign optimal weights to sub-classifiers.

\subsection{Online Classification with Limited Supervision}
Since the fully-supervised data is hard and expensive to collect in real-world applications, many works is proposed to study different settings of online classification with limited supervision. The state of the arts of limited-supervised online classification can be categorized into multi-class methods and multi-label methods according to the data type to be classified. For multi-class methods, Wang \emph{et.al} \cite{OPLL} proposed and solved online partial label learning where each data example is associated with multiple candidate labels. To further handle the possibility that new classes emerge in open and dynamic environment, the work of \cite{OPLLE} incorporates an ensemble-based detector to identify new classes and label disambiguation process to tackle candidate labels. The work of \cite{Adaptive} considers a more general noisy case that the instances belonging to one class may be assigned to another class, and acts stream classification based on the estimated noise transition matrix. For multi-label methods, Li \emph{et.al} \cite{semi1} addressed semi-supervised online multi-label classification via label compression and local smooth regression, while the work of \cite{semi2} integrates the kernel function into the well-developed ELM framework for online semi-supervised classification. To adapt to the environment with missing labels, Bakhshi \emph{et.al} \cite{Balance} uses a simple imputation strategy with a selective concept drift adaptation mechanism to well suit concept drift, and Wang \emph{et al.} \cite{OMAL2} builts an ensemble-based active learning framework for the consideration of labelling cost.

The online methods mentioned above all have limitations on the quality and type of the label set while the more general noisy multi-label case remain unexplored, which assumes noisy labels can occur in both relevant and irrelevant labels. Our method not only addresses the problem of online noisy multi-label classification, but also considers the possible label distribution changes in online environment.

\section{Our method}
In this section, the newly developed objective function based on the label scoring and ranking framework is proposed to deal with the noisy label problem, whose closed-form solution indicates the efficient model update and high model performance. Next, the sequential update rules are derived in detail. Finally, to handle two cases of noisy and changing label distributions, we derive an unbiased statistic with respect to label cardinality to detect the potential concept drift, and propose two novel strategies to adapt to it.

\subsection{Problem Formulation}
Firstly we give the problem formulation of OMC with noisy labels. Given a multi-label data stream $D = [D_{i}], i = 1,2,...$ with potentially infinite number of data chunks, each chunk $D_{i} = \{\mathbf{X}_i,\mathbf{Y}_i\}$ equally contain $N$ data instances. Within it, $\mathbf{X}_i = [\mathbf{x}^i_{t}], t \in [N]$ indicates the feature space in $D_{i}$ ($[N] = \{1,...,N\}$) and $\mathbf{Y}_i = [\mathbf{y}^i_{t}], t \in [N]$ represents the observed label space in $D_{i}$ with noisy labels, where $\mathbf{x}^i_{t} \in \mathbb{R}^d,\mathbf{y}^i_{t} \in \mathbb{R}^q$ denote the $d$-dimensional feature vector and $q$-dimensional label vector of the $t$-th instance in $D_{i}$ respectively. In data chunk $D_{i}$, if the $j$-th label of the $t$-th instance is tagged as relevant, $y^i_{tj} = 1$, conversely $y^i_{tj} = -1$. Correspondingly, we denote the ground-truth label matrix as $\mathbf{G}_i = [\mathbf{g}^i_{t}], t \in [N]$, which is unavailable in the process of online classification. Instead, the noisy rates throughout the online data with respect to each class label are known: 

\begin{equation}
\begin{aligned}
\label{NR}
&p\left(y^i_{tj}=-1 \mid g^i_{tj}=+1\right)=\rho_{+}^j, \\ &p\left(y^i_{tj}=+1 \mid g^i_{tj}=-1\right)=\rho_{-}^j, \\ &\forall i \geq 1, j \in [q], \rho_{+}^j+\rho_{-}^j<1 .
\end{aligned} 
\end{equation}
Where $\rho_{+}^j, \rho_{-}^j$ denote the rates that relevant (irrelevant) $j$-th labels are flipped into irrelevant (relevant) labels. 

Given the data stream $D$ with the noisy rate defined above, when 
$D_i$ arrives at time $i$, the task of OMC with noisy labels is to utilize the model $\mathbf{\Phi}_{i-1}$ learned from past data $[D_{u}], u \in [i-1]$ to score the labels of each instance in $D_i$ as $o^i_{tj}$, thus, as mentioned in Section \uppercase\expandafter{\romannumeral 1}, the relevant label set of the $t$-th instance is predicted as $\hat{\mathbf{y}}^i_t = \{\hat{y}^i_{tj}|o^i_{tj} > 0, j \in [q]\}$, then $\mathbf{Y}_i$ is observed and the $\mathbf{\Phi}_{i}$ is obtained based on the sequential update rule. The details of performing robust OMC with noisy labels are given in the following section \uppercase\expandafter{\romannumeral 3}-B.

Additionally, considering the concept drift in the ground-truth label distribution, we formulate two cases of NCLD as: 1) the ground-truth concept growth $D^S_G \rightarrow D^M_G$; 2) the ground-truth concept reduction $D^M_G \rightarrow D^S_G$, where $D^S_G, D^M_G$ denote the data streams with ground-truth single-label and multi-label distributions, respectively. The task of OMC under NCLD is to detect and adapt to the above two cases of ground-truth concept drift only with the noisy observation $D$, thus achieving the robustness to both noisy labels and varying distributions (NCLD), the corresponding details are presented in Section \uppercase\expandafter{\romannumeral 3}-D. 

Using $L$ random input weights and biases $\mathbf{w}_i, \mathbf{b}_i, i \in [L]$ and the \emph{sigmoid} function as the activation function, the output matrix of data chunk $\mathbf{X}_i$ with respect to hidden layer is formulated as follow: 

\begin{equation}
\begin{aligned}
\label{ELM}
\mathbf{H}_i=\left[\begin{array}{ccc}\sigma\left(\boldsymbol{\alpha}_1, b_1, \mathbf{x}^i_1\right) & \cdots & \sigma\left(\boldsymbol{\alpha}_L, b_L, \mathbf{x}^i_1\right) \\ \vdots & \ddots & \vdots \\ \sigma\left(\boldsymbol{\alpha}_1, b_1, \mathbf{x}^i_{N}\right) & \cdots & \sigma\left(\boldsymbol{\alpha}_L, b_L, \mathbf{x}^i_{N}\right)\end{array}\right]_{N \times L} 
\end{aligned} 
\end{equation}

Instead of classifying with $\mathbf{X}_i$ using the original $d$ features, we score the labels of the instances in $D_i$ using the transformed $\mathbf{H}_i$ to improve efficiency ($L \ll d$). 

\subsection{OMC with Noisy Labels}
Without considering the concept drift, we first formulate the objective function robust to noisy labels. As mentioned in Section \uppercase\expandafter{\romannumeral 1}, two parts are incorporated
in the objective including the label scoring and label ranking term. With the online model $\mathbf{\Phi}_{i-1}$ trained based on $[D_{u}], u \in [i-1]$, the label scores of the \emph{t}-th instance in $D_{i}$ is computed as $\mathbf{o}^i_t = \mathbf{h}^i_t\mathbf{\Phi}_{i-1} \in \mathbb{R}^q$ where $\mathbf{h}^i_t$ is obtained with $\mathbf{x}^i_t$ via ELM. The \emph{j}-th component $o^i_{tj}$ of $\mathbf{o}^i_t$ indicates the score with respect to the \emph{j}-th label. Usually for noiseless data, the mean square error is formulated as:

\begin{equation}
\begin{aligned}
\label{N1}
\underset{\mathbf{\Phi} \in \mathbb{R}^{L\times q}} {\operatorname{min}} &\frac{1}{2}\|\mathbf{O}-\mathbf{Y}\|_F^2 + \frac{\alpha}{2} \|\mathbf{\Phi}  \|_F^2 \\
&\text { s.t. } \ \mathbf{O} = \mathbf{H}\mathbf{\Phi}
\end{aligned} 
\end{equation}
where $\mathbf{O} \in \mathbb{R}^{N \times q}$ denotes the scoring matrix and each element in the \emph{t}-th row and the \emph{j}-th column  corresponds to each $o^i_{tj}$ with respect to the \emph{i}-th chunk (for simplicity, below we abbreviate the symbol $i$ without confusion), the second term gives the $L_2$-regularization constraint on the model coefficient matrix $\mathbf{\Phi} \in \mathbb{R}^{L\times q}$ and $\alpha$ is the regularization factor. However, the first term indicates the optimal scores of each instance-label pair $o^*_{tj} = y_{tj}$, which leads to the inferior solution due to the potential noisy label $y_{tj}$. 

To handle noisy labels, we utilize the relation between each instance regarding feature space to reconstruct the label scores. Firstly, we construct a weighted directed graph $\mathcal{G}$, which is instantiated using \emph{K}-Nearest-Neighbours error minimization based on the feature space $\mathbf{X}^i = [\mathbf{x}^i_t], t \in [N]$ in the current chunk \emph{i}:
\begin{equation}
\begin{aligned}
\label{Re}
\min _{\mathbf{S} \in 
\mathbb{R}^{N\times N} }& \left\|\mathbf{X} - \mathbf{S}\mathbf{X}\right\|_{F}^{2}  \\
\text { s.t. } & \mathbf{S} \succeq 0, \ \mathbf{S}\mathbf{1_N} = \mathbf{1_N},\\
&  S_{n,n} = 0, S_{n,m} = 0 \quad\left(m \notin \mathcal{N}\left({n}\right)\right)
\end{aligned}
\end{equation}

Here, the graph $\mathcal{G}$ is locally instantiated on $D_i$ by viewing each instance in $D_i$ as a point and $S_{n,m}$ indicates the weight from point $m$ to point $n$. Based on the reconstruction weights information $\mathbf{S}$ learnt on the feature space, we reconstruct the label scores of each instance both from the observed label and the scores of its nearest neighbours:

\begin{equation}
\begin{aligned}
\label{N2}
\underset{\mathbf{\Phi} \in \mathbb{R}^{L\times q}}{\operatorname{min}} &\frac{\beta}{2}\|\mathbf{O}-\mathbf{Y}\|_F^2 + \frac{1-\beta}{2}
\| \mathbf{O} - \mathbf{S}\mathbf{O}    \|_F^2 + \frac{\alpha}{2} \|\mathbf{\Phi}\|_F^2\\
&\text { s.t. } \ \mathbf{O} = \mathbf{H}\mathbf{\Phi}
\end{aligned} 
\end{equation}
where $\beta \in [0,1]$ is a weight factor. We have the following proposition:

\noindent \textbf{Proposition 1.} The optimal label score $o^*_{tj}$ optimized by the Eq. \eqref{N2} meets the following equation:\\
\begin{equation}
\label{P1}
o^*_{tj} =  \beta y^*_{tj} + (1- \beta) \sum_{n} S_{t,n}o^*_{nj}, n \in \mathcal{N}\left({t}\right) 
\end{equation}

With Eq. \eqref{N2}, proposition 1 demonstrates the score of each instance-label pair $o_{tj}$ can be reconstructed using the weighted sum of the label $y_{tj}$ and the scores of the nearest neighbours of the \emph{t}-th instance. Note that $\beta = 1$ indicates the case where only $y_{tj}$ is used to reconstruct $o_{tj}$, namely $o^*_{tj} = y_{tj}$ in accordance with Eq. \eqref{N1}. Thus, by setting a $\beta \in (0,1)$, we can control the extent to which nearest neighbour scores are used to improve model robustness. By far, Eq. \eqref{N2} has incorporated the robust label scoring term ensuring that the scores of relevant labels and irrelevant labels can follow the bipartition over the fixed threshold of 0. Now we need to further correctly rank the scores between relevant labels and irrelevant labels.

Given the data stream $D_G$ with the ground-truth label distribution (i.e., $\mathbf{y}_j=\mathbf{g}_j$), we aim to minimize the \emph{ranking loss} of classifying each stream instance with respect to all label pairs $(\mathbf{y}_j,\mathbf{y}_k), j,k \in [q]$:

\begin{equation}
\begin{aligned}
\label{RL}
\underset{\mathbf{\Phi} \in \mathbb{R}^{L\times q}}{\operatorname{min}} \sum_{t} \sum_{j=1}^q \sum_{k=1}^q
f(y^{(t)}_{j,k}o^{(t)}_{j,k})
\ \text { s.t. } \ \mathbf{O} = \mathbf{H}\mathbf{\Phi}
\end{aligned}
\end{equation}
where $y^{(t)}_{j,k} = (y_{tj}-y_{tk})/2$, $o^{(t)}_{j,k} = o_{tj}-o_{tk}$, $f(\cdot)$ is a convex loss function and the gradient of $f(\cdot)$ with respect to $o^{(t)}_{j,k}$ is equal to 0 given $y^{(t)}_{j,k}=0$, which is consistent with the fact that only the relevant-irrelevant label pairs form the loss.

Taking into account the impact led by the noisy labels, the unbiased estimator with respect to $f(y^{(t)}_{j,k}o^{(t)}_{j,k})$ under the ground-truth label distribution is formulated as the following proposition: 

\noindent \textbf{Proposition 2.} Considering the feature, \emph{j}-th label and \emph{k}-th label of the \emph{t}-th instance namely $\mathbf{x}_t,y_{tj},y_{tk}$ respectively, $D_G,D$ denote the ground-truth and noisy data distribution, the following equality holds: 

\begin{equation}
\begin{aligned}
\label{RL_Derive}
\mathbb{E}_{(\mathbf{x}_t,y_{tj},y_{tk}) \sim D_{G}} &(f(y^{(t)}_{j,k}o^{(t)}_{j,k}))  = \\ &
\mathbb{E}_{(\mathbf{x}_t,y_{tj},y_{tk}) \sim D} (\omega_{t,j}\omega_{t,k}f(y^{(t)}_{j,k}o^{(t)}_{j,k}))
\end{aligned}
\end{equation}
where $\omega_{t,j} = \frac{P_{D_G}(y_{tj} \mid \mathbf{x}_t)}{P_{D}(y_{tj} \mid \mathbf{x}_t)}, \omega_{t,k} = \frac{P_{D_G}(y_{tk} \mid \mathbf{x}_t)}{P_{D}(y_{tk} \mid \mathbf{x}_t)}$. Given the noisy ratio $\rho_{+}^j, \rho_{-}^j$ with respect to the label $j$, the following equality holds \cite{Reweighting}:

\begin{equation}
\begin{aligned}
\omega_{t,j} = \frac{P_{D}\left(y_{tj} \mid \mathbf{x}_t\right)-\rho_{-y_{tj}}}{\left(1-\rho_{+}^j-\rho_{-}^j\right) P_{D}\left(y_{tj} \mid \mathbf{x}_t\right)}
\end{aligned}
\end{equation}

To compute $\omega_{t,j}$, it is necessary to estimate the probability output $P_{D}\left(y_{tj} \mid \mathbf{x}_t\right)$. Specifically, with respect to the \emph{j}-th label of the \emph{t}-th instance in the noisy data chunk $D_i$, we compute its sigmoid output based on the ELM model trained separately on each $D_i$ to estimate the probability output, which fits local distribution of each $D_i$ to account for possible potential concept drift. There are many options for the loss function 
$f(y^{(t)}_{j,k}o^{(t)}_{j,k})$ such as the square loss $f(x) = (1-x)^2$ and the hinge loss $f(x) = (1-x)_+$, while we choose the affine function $f(x) = -x$ for a feasible and efficient online sequential update as shown in section \uppercase\expandafter{\romannumeral 3}-C. Thus, given the data stream $D$ with noisy labels, we rewrite the unbiased form of Eq. \eqref{RL} under the noiseless distribution $D_G$ and substitute $f(x) = -x$ into it:

\begin{equation}
\begin{aligned}
\label{RL_New}
\underset{\mathbf{\Phi} \in \mathbb{R}^{L\times q}}{\operatorname{min}} \sum_{t} \sum_{j=1}^q \sum_{k=1}^q
\omega_{t,j}\omega_{t,k}y^{(t)}_{k,j}(o_{tj}-o_{tk})
\ \text { s.t. } \ \mathbf{O} = \mathbf{H}\mathbf{\Phi}
\end{aligned}
\end{equation}

By further integrating the unbiased ranking loss term into Eq. \eqref{N2}, we achieve the final objective:
\begin{equation}
\begin{aligned}
\label{N3}
\underset{\mathbf{\Phi} \in \mathbb{R}^{L\times q}}{\operatorname{min}} &\frac{\beta}{2}\|\mathbf{O}-\mathbf{Y}\|_F^2 + \frac{1-\beta}{2}
\|\mathbf{O} - \mathbf{S}\mathbf{O}    \|_F^2 \\
&+ \gamma \text{Tr}(\textbf{A}^\mathrm{T}\mathbf{O}) + \frac{\alpha}{2} \|\mathbf{\Phi}\|_F^2 \\
&\text { s.t. } \ \mathbf{O} = \mathbf{H}\mathbf{\Phi}
\end{aligned} 
\end{equation}
where $\gamma$ is the regularization factor of the label ranking term. Within the matrix $\mathbf{A} \in \mathbb{R}^{N \times q}$, each element $A_{t,j} = \omega_{t,j}\sum_{k=1}^q \omega_{t,k}y^{(t)}_{k,j}$. Then, the gradient of Eq. \eqref{N3} with respect to $\mathbf{\Phi}$ is:
\begin{equation}
\begin{aligned}
\label{Gra}
\nabla_{\mathbf{\Phi}} &= \left[\alpha\mathbf{I} + \mathbf{H}^\mathrm{T}\big(\beta\mathbf{I}+(1-\beta)(\mathbf{S}^\mathrm{T}\mathbf{S}-\mathbf{S}^\mathrm{T}-\mathbf{S})\big)\mathbf{H}\right]\mathbf{\Phi}  \\ &
- \mathbf{H}^\mathrm{T}(\beta\mathbf{Y}-\gamma\mathbf{A})
\end{aligned} 
\end{equation}
where $\mathbf{I} \in \mathbb{R}^{d \times d}$ is the identity matrix. By setting $\nabla_{\mathbf{\Phi}}=0$, we have the closed-form solution of Eq. \eqref{N3}:
\begin{equation}
\begin{aligned}
\label{Gra0}
\mathbf{\Phi}^* &= (\alpha\mathbf{I}+\mathbf{H}^\mathrm{T}\mathbf{R}  \mathbf{H})^{-1}\mathbf{H}^\mathrm{T}(\beta\mathbf{Y}-\gamma\mathbf{A})
\end{aligned} 
\end{equation}
where $\mathbf{R} = \beta\mathbf{I}+(1-\beta)(\mathbf{S}^\mathrm{T}\mathbf{S}-\mathbf{S}^\mathrm{T}-\mathbf{S}$). Hence in objective shown in Eq. \eqref{N3}, the regularization term can help to effectively integrate label scoring and ranking information to build the OMC model robust to noisy labels.

\subsection{Sequential Update Rule}
\emph{1) Initialization Phase:} Before online classification, it is common to train a base model $\mathbf{\Phi}_0$ on the initialization chunk $D_0 = \{\mathbf{X}_0, \mathbf{Y}_0\}$. According to Eq. \eqref{Gra0}, the optimal solution meets:
\begin{equation}
\begin{aligned}
\label{Phi_0}
\mathbf{\Phi}_0 &= \mathbf{K}_0^{-1}\mathbf{H}_0^\mathrm{T}\mathbf{M}_0
\end{aligned} 
\end{equation}
where $\mathbf{K}_0 = \alpha\mathbf{I}+\mathbf{H}_0^\mathrm{T}\mathbf{R}_0  \mathbf{H}_0 \in \mathbb{R}^{d \times d}, \mathbf{M}_0 = \beta\mathbf{Y}_0-\gamma\mathbf{A}_0 \in \mathbb{R}^{N \times q}$, , $\mathbf{H}_0$ is calculated by Eq. \eqref{ELM}, and the definitions of $\mathbf{R}_0$ and $\mathbf{A}_0$ have been given in the Section-\uppercase\expandafter{\romannumeral 3}-B.

\emph{2) Sequential Learning Phase:} After initializing the model 
$\mathbf{\Phi}_0$ based on $D_0$, we need to update the model as $\mathbf{\Phi}_1$ using the newly incoming data chunk $D_1$, surely $\mathbf{\Phi}_1$ satisfy the following:
\begin{equation}
\begin{aligned}
\mathbf{\Phi}_1 &= \mathbf{K}_1^{-1}\left[\begin{array}{c} \mathbf{H}_{0} \\ \mathbf{H}_{1}\end{array}\right]^\mathrm{T}\left[\begin{array}{c} \mathbf{M}_{0} \\ \mathbf{M}_{1}\end{array}\right]
\end{aligned} 
\end{equation}

Since the weight $\mathbf{S}$ of the graph is localized on each $D_i$, we have:

\begin{equation}
\begin{aligned}
\label{K_Update}
\mathbf{K}_1 = & \ \alpha\mathbf{I} + \left[\begin{array}{c} \mathbf{H}_{0} \\ \mathbf{H}_{1}\end{array}\right]^\mathrm{T} 
\Bigg( \left[\begin{array}{cc}
\mathbf{S}_{0} &\mathbf{0} \\
\mathbf{0} & \mathbf{S}_{1}
\end{array}\right]^\mathrm{T}
\left[\begin{array}{cc}
\mathbf{S}_{0}  &\mathbf{0} \\
\mathbf{0} & \mathbf{S}_{1}
\end{array}\right] - \\
& \left[\begin{array}{cc}
\mathbf{S}_{0} &\mathbf{0} \\
\mathbf{0} & \mathbf{S}_{1}
\end{array}\right]^\mathrm{T} - 
\left[\begin{array}{cc}
\mathbf{S}_{0} &\mathbf{0} \\
\mathbf{0} & \mathbf{S}_{1}
\end{array}\right] \Bigg) \left[\begin{array}{c} \mathbf{H}_{0} \\ \mathbf{H}_{1}\end{array}\right] \\
& =  \alpha\mathbf{I} + \mathbf{H}_0^\mathrm{T}\mathbf{R}_0  \mathbf{H}_0 +  \mathbf{H}_1^\mathrm{T}\mathbf{R}_1 \mathbf{H}_1 \\
& = \mathbf{K}_0 + \mathbf{H}_1^\mathrm{T}\mathbf{R}_1  \mathbf{H}_1
\end{aligned}
\end{equation}
where $\mathbf{R}_1 = \beta\mathbf{I}+(1-\beta)(\mathbf{S}_1^\mathrm{T}\mathbf{S}_1-\mathbf{S}_1^\mathrm{T}-\mathbf{S}_1)$, and the last equality follows from $\mathbf{K}_0 = \alpha\mathbf{I} + \mathbf{H}_0^\mathrm{T}\mathbf{R}_0  \mathbf{H}_0$. In the current round, the reverse of $\mathbf{K}_1$ can be obtained via the Woodbury formula \cite{matrix}:
\begin{equation}
\begin{aligned}
\label{P_Update}
\mathbf{K}_1^{-1} &= (\mathbf{K}_0 + \mathbf{H}_1^\mathrm{T}\mathbf{R}_1\mathbf{H}_1)^{-1} 
\\ & = \mathbf{K}_0^{-1} - \mathbf{K}_0^{-1}\mathbf{H}_1^\mathrm{T}(\mathbf{R}_1^{-1}+\mathbf{H}_1\mathbf{K}_0^{-1}\mathbf{H}_1^\mathrm{T})^{-1}\mathbf{H}_1\mathbf{K}_0^{-1}
\end{aligned}
\end{equation}

Thus we can compute $\mathbf{K}_1^{-1}$ using $\mathbf{K}_0^{-1}$, next, we can derive the sequential update rule for $\mathbf{\Phi}_1$: 
\begin{equation}
\begin{aligned}
\label{Phi_Update}
\mathbf{\Phi}_1 &= \mathbf{K}_1^{-1}\left[\begin{array}{c} \mathbf{H}_{0} \\ \mathbf{H}_{1}\end{array}\right]^\mathrm{T}\left[\begin{array}{c} \mathbf{M}_{0} \\ \mathbf{M}_{1}\end{array}\right] 
 \\ & = \mathbf{K}_1^{-1} (\mathbf{H}_{0}^\mathrm{T}\mathbf{M}_{0}+\mathbf{H}_{1}^\mathrm{T}\mathbf{M}_{1})
 \\ & = \mathbf{K}_1^{-1}(\mathbf{K}_0\mathbf{\Phi}_0+\mathbf{H}_{1}^\mathrm{T}\mathbf{M}_{1})
\\ & = \mathbf{K}_1^{-1}\big((\mathbf{K}_1-\mathbf{H}_1^\mathrm{T}\mathbf{R}_1  \mathbf{H}_1) \mathbf{\Phi}_0+\mathbf{H}_{1}^\mathrm{T}\mathbf{M}_{1}\big)
\\ & = \mathbf{\Phi}_0 - \mathbf{K}_1^{-1}(\mathbf{H}_1^\mathrm{T}\mathbf{R}_1\mathbf{H}_{1}\mathbf{\Phi}_0 - \mathbf{H}_{1}^\mathrm{T}\mathbf{M}_{1})
\end{aligned}
\end{equation}
the third and fourth equality hold based on Eq. \eqref{Phi_0} and Eq. \eqref{K_Update}, respectively. After denoting $\mathbf{P}_i = \mathbf{K}_i^{-1}$ for notation simplicity, we obtain the following update rule based on Eq. \eqref{P_Update} and Eq. \eqref{Phi_Update}:
\begin{equation}
\begin{aligned}
\label{Update}
\mathbf{P}_1 &=  \mathbf{P}_0 - \mathbf{P}_0\mathbf{H}_1^\mathrm{T}(\mathbf{R}_1^{-1}+\mathbf{H}_1\mathbf{P}_0\mathbf{H}_1^\mathrm{T})^{-1}\mathbf{H}_1\mathbf{P}_0
 \\ \mathbf{\Phi}_1 & = \mathbf{\Phi}_0 - \mathbf{P}_1(\mathbf{H}_1^\mathrm{T}\mathbf{R}_1\mathbf{H}_{1}\mathbf{\Phi}_0 - \mathbf{H}_{1}^\mathrm{T}\mathbf{M}_{1})
\end{aligned}
\end{equation}

As the (\emph{i}+1)-th chunk $D_{i+1}$ arrives, we generalize the above equation to give the recursive formula for the online classification model:
\begin{equation}
\begin{aligned}
\label{Update}
\mathbf{P}_{i+1} &=  \mathbf{P}_{i} - \mathbf{P}_{i}\mathbf{H}_{i+1}^\mathrm{T}(\mathbf{R}_{i+1}^{-1}+\mathbf{H}_{i+1}\mathbf{P}_{i}\mathbf{H}_{i+1}^\mathrm{T})^{-1}\mathbf{H}_{i+1}\mathbf{P}_{i}
 \\ \mathbf{\Phi}_{i+1} & = \mathbf{\Phi}_{i} - \mathbf{P}_{i+1}(\mathbf{H}_{i+1}^\mathrm{T}\mathbf{R}_{i+1}\mathbf{H}_{i+1}\mathbf{\Phi}_{i} - \mathbf{H}_{i+1}^\mathrm{T}\mathbf{M}_{i+1})
\end{aligned}
\end{equation}

Based on the above update rule derived above, the model $\mathbf{\Phi}_i$ that we update at time \emph{i} is equivalent to the model trained on $[D_{u}], u \in [i]$ as a batch. In addition, the time taken for each update is limited by the number of hidden layer nodes $L \ll d$. Hence, the closed-form solution optimized based on the data ever seen is efficiently updated at each round, which can effectively achieve competitive performance compared to batch and fully-supervised methods. 

\subsection{OMC under NCLD}
As mentioned in Section \uppercase\expandafter{\romannumeral 3}-A, we consider two types of NCLD here: 1) the ground-truth concept growth $D^S_G \rightarrow D^M_G$; 2) the ground-truth concept reduction $D^M_G \rightarrow D^S_G$, $D^S_G$ and $D^M_G$ denote the stream data with the single-label and multi-label distributions respectively. With the robustness of the model to noisy labels well-established, the remaining problem is about how to detect the ground-truth concept drift given the noisy data stream $D$. Based on the intuition that the ground-truth cardinality \cite{Review} of each data chunk can be used to track the concept drift ($LCard(D_i) = \frac{1}{N}\sum_{t=1}^N \sum_{j=1}^q \mathbb{I}_{\{ g_{tj} = 1 \}}$), the following proposition is proposed to give an unbiased estimation of the ground-truth cardinality given the noisy observation $D_i = \{\mathbf{X}_i,\mathbf{Y}_i\}$: 

\noindent \textbf{Proposition 3.} Given the noisy and the ground-truth stream data distribution denoted as $D$ and $D_G$, the following equality w.r.t. the \emph{t}-th instance holds: 

\begin{equation}
\begin{aligned}
\label{card}
&\mathbb{E}_{(\mathbf{x}_t,\mathbf{y}_{t}) \sim D_{G}}\big(LCard(\mathbf{x}_t,\mathbf{y}_{t})\big) =
\sum_{j=1}^q \mathbb{E}_{(\mathbf{x}_t,y_{tj}) \sim D} \left(\mathbb{I}_{\{ y_{tj} = 1 \}} \omega_{t,j}\right)
\\ & =
\sum_{j=1}^q \mathbb{E}_{(\mathbf{x}_t,y_{tj}) \sim D}  \left( \mathbb{I}_{\{ y_{tj} = 1 \}} \frac{P_{D}\left(y_{tj} \mid \mathbf{x}_t\right)-\rho_{-y_{tj}}}{(1-\rho_{+}^j-\rho_{-}^j)P_{D}\left(y_{tj} \mid \mathbf{x}_t\right)}\right)
\end{aligned}
\end{equation}
where $\mathbb{I}(\cdot)$ is the indicator function, $\mathbb{I}(x) = 1$ if the predicate $x$ holds, conversely $\mathbb{I}(x) = 0$.

Therefore, the cardinality of the data chunk $D_i$
can be reliably estimated via the the empirical form of Eq. \eqref{card}:
$\hat{LCard}(D_i) = 
\frac{1}{N}\sum_{t=1}^N \sum_{j=1}^q \mathbb{I}_{\{ y_{tj} = 1 \}} \omega_{t,j}$. We use the cardinality difference between two adjacent chunks to detect the potential concept drift, since not only concept growth but also concept reduction will lead to a significant cardinality difference. In other words, if $|\hat{LCard}(D_{i}) - \hat{LCard}(D_{i-1})| > \varepsilon_{i}$ holds when the concept drift occurs on the data chunk $D_i$, where the following proposition computes $\varepsilon_{i}$ based on the Hoeffding inequality \cite{hfd}:

\noindent \textbf{Proposition 4.} Given the label cardinality estimation $[\hat{LCard}(\mathbf{x}^i_t)], t \in [N]$ in the current data chunk $D_i$, the threshold with respect to the $i$-th round can be calculated as:
\begin{equation}
\begin{aligned}
\label{Hfd}
\varepsilon_i = R_i\sqrt{\frac{ln(2/\delta)}{2N}}
\end{aligned}
\end{equation}
where $R_i = \max_{t}[\hat{LCard}(\mathbf{x}^i_t)]-\min_{t}[\hat{LCard}(\mathbf{x}^i_t)], t \in [N]$ and $\delta \in (0,1)$ denotes the confidence value.

Therefore, if we find that $|\hat{LCard}(D_{i}) - \hat{LCard}(D_{i-1})| > \varepsilon_{i}$, we can conclude that the ground truth concept drift is occurring on the \emph{i}-th data chunk. Once the concept drift is detected, we need to update the model for drift adaption \cite{Review_CD}. The following two adaption methods are provided:

\emph{1) Retrain the model:} Since the data distribution changes from one to the other, the most straightforward strategy is to abandon the past model established on the old data $[D_u], u \in [i-1]$ and retrain the model on the new data $D_i$. Thus, it can be achieved by simply setting $\mathbf{\Phi}_{i-1} = \mathbf{0}$ when the concept drift is detected on $D_i$, then initializing $\mathbf{P}_{i}$ only with $D_i$ and retraining the new model $\mathbf{\Phi}_{i}$ based on $D_i$, $\mathbf{P}_{i}$ and the reset $\mathbf{\Phi}_{i-1}$ using Eq. \eqref{Update}. However, this update strategy may be too aggressive to retain valuable information learned from the past. 

\emph{2) Adjust the model:} The label scoring information learned by $\mathbf{\Phi}_{i-1}$ is likely to promote the classification performance for the new data distribution, thus we can only reset the old label ranking information of the model as an adjustment when the concept drift occurs. Specifically, the optimal solution with respect to the Eq. \eqref{N2} is denoted as $\mathbf{\Psi}$, which only takes into account  the label scoring information. The basic idea is that when concept drift is detected, the model $\mathbf{\Phi}_{i-1}$ derived from Eq. \eqref{N3} is replaced by $\mathbf{\Psi}_{i-1}$ without incorporating the old label ranking information, then the new $\mathbf{\Phi}_{i}$ is updated leveraging the old label scoring information from $\mathbf{\Phi}_{i-1}$ as well as new label scoring and ranking information from $D_i$.

It is necessary to calculate $\mathbf{\Psi}_{i}$ when concept drift occurs on $D_i$, thus we propose to calculate the following 
matrix sequence $[\mathbf{Z}_i] = [\mathbf{\Psi}_{i} - \mathbf{\Phi}_{i}]$, and $\mathbf{\Psi}_{i}$ can be calculated as $\mathbf{\Psi}_{i} = \mathbf{\Phi}_{i} + \mathbf{Z}_i$. The relation between $\mathbf{Z}_{1}$ and $\mathbf{Z}_{0}$ is computed as:
\begin{equation}
\begin{aligned}
\label{Z1_Update}
\mathbf{Z}_1 &= \mathbf{\Psi}_{1} - \mathbf{\Phi}_{1} 
\\ & = \mathbf{K}_{1}^{-1}(\gamma\mathbf{H}_1^\mathrm{T}\mathbf{A}_1+\gamma\mathbf{H}_0^\mathrm{T}\mathbf{A}_0)
\\ & = \mathbf{K}_{1}^{-1}(\gamma\mathbf{H}_1^\mathrm{T}\mathbf{A}_1+\mathbf{K}_{0}\mathbf{Z}_0)
\\ & = \mathbf{K}_{1}^{-1}\left(\gamma\mathbf{H}_1^\mathrm{T}\mathbf{A}_1+(\mathbf{K}_{1}-\mathbf{H}_1^\mathrm{T}\mathbf{R}_1\mathbf{H}_1)\mathbf{Z}_0\right) 
\\ & = 
\mathbf{Z}_0 - \mathbf{K}_{1}^{-1}(\mathbf{H}_1^\mathrm{T}\mathbf{R}_1\mathbf{H}_1 \mathbf{Z}_0 - \gamma\mathbf{H}_1^\mathrm{T}\mathbf{A}_1)
\\ & = 
\mathbf{Z}_0 - \mathbf{P}_{1}(\mathbf{H}_1^\mathrm{T}\mathbf{R}_1\mathbf{H}_1 \mathbf{Z}_0 - \gamma\mathbf{H}_1^\mathrm{T}\mathbf{A}_1)
\end{aligned}
\end{equation}
and the final update rule is:
\begin{equation}
\begin{aligned}
\label{Z_Update}
\mathbf{Z}_{i+1} & = \mathbf{Z}_{i} - \mathbf{P}_{i+1}(\mathbf{H}_{i+1}^\mathrm{T}\mathbf{R}_{i+1}\mathbf{H}_{i+1}\mathbf{Z}_{i} - \gamma\mathbf{H}_{i+1}^\mathrm{T}\mathbf{A}_{i+1})
\end{aligned}
\end{equation}

Note that the $\mathbf{P}_{i+1}$ can be updated once each round via Eq. \eqref{Update} for both updating $\mathbf{\Phi}_{i+1}$ and $\mathbf{Z}_{i+1}$, leading to efficient updating of $\mathbf{Z}_{i+1}$. 
Once concept drift occurs, the model is set as $\mathbf{\Phi}_{i-1} = \mathbf{\Psi}_{i-1} = \mathbf{\Phi}_{i-1}+\mathbf{Z}_{i-1}$ and $\mathbf{Z}_{i-1}$ is reset as zero matrix, then the new $\mathbf{\Phi}_{i}, \mathbf{Z}_{i}$ are updated via Eq. \eqref{Update} and Eq. \eqref{Z_Update}. By incorporating the detection and adaptation methods to account for concept drift hidden in the noisy data, our ELM-based model is able to effectively and efficiently classify multi-label stream data under noisy and changing label distribution (ELM-NCLD), whose overall framework is summarized in Algorithm 1.
\begin{algorithm}[t]
    \caption{The algorithm of ELM-NCLD}
    \label{Algorithm1}
    \begin{algorithmic}[1]
    \REQUIRE  
    The initialization data chunk $D_0$, the data stream with noisy labels $[D_i], i = 1,2,...$;
    Regularization factors $\alpha, \beta, \gamma$; Other parameters 
    $\delta,N,L$;
    \ENSURE $\mathbf{\Phi}_{i}$ at each round \emph{i};
    \STATE// Initialize the model based on $D_0$
    \STATE Initialize $\mathbf{P}_0 = (\alpha\mathbf{I}+\mathbf{H}_0^\mathrm{T}\mathbf{R}_0\mathbf{H}_0)^{-1}, \mathbf{\Phi}_{0} = \mathbf{P}_0\mathbf{H}_0^\mathrm{T}\mathbf{M}_0$, 
   $\mathbf{Z}_{0} = \gamma\mathbf{P}_0\mathbf{H}_0^\mathrm{T}\mathbf{A}_0$.\\ 
    \STATE// Start online sequential classification
    \WHILE{remain available data}
    \STATE Receive the \emph{i}-th data chunk $D_i=\{(\mathbf{X}_i,\mathbf{Y}_i)\}$; 
    \STATE Transform $\mathbf{X}_i$ into 
    $\mathbf{H}_i$ via Eq. \eqref{ELM};
    \STATE Compute the label score $\mathbf{o}^i_{t} = \mathbf{h}_t\mathbf{\Phi}_{i-1}, t \in [N]$;
    \STATE Predict relevant labels as  $\hat{\mathbf{y}}^i_t = \{\hat{y}^i_{tj}|o^i_{tj} > 0, j \in [q]\}$;
    \STATE Calculate $\mathbf{S}_i$ via Eq. \eqref{Re} and $\mathbf{R}_i, \mathbf{M}_i$ via Eq. \eqref{Phi_0};
    \STATE Update $\mathbf{P}_{i}$ via Eq. \eqref{Update};
    \STATE// Detect and adapt to concept drift 
    \STATE Compute $\hat{LCard}(D_i), R_i$ via Eq. \eqref{card};
    \STATE Compute $\varepsilon_{i}$ via Eq. \eqref{Hfd};
    \IF{$|\hat{LCard}(D_{i}) - \hat{LCard}(D_{i-1})| > \varepsilon_{i}$}
    \IF{\emph{strategy} == '\emph{retrain}'}
    \STATE $\mathbf{\Phi}_{i-1} = \mathbf{0}, \mathbf{P}_{i} = (\alpha\mathbf{I}+\mathbf{H}_i^\mathrm{T}\mathbf{R}_i\mathbf{H}_i)^{-1}$;
    \ELSE
    \STATE $\mathbf{\Phi}_{i-1} = \mathbf{\Phi}_{i-1}+\mathbf{Z}_{i-1},\mathbf{Z}_{i-1} = \mathbf{0}$;
    \ENDIF
    \ENDIF
    \STATE Update $\mathbf{\Phi}_{i},\mathbf{Z}_{i}$ via Eq. \eqref{Update} and Eq. \eqref{Z_Update}; 
    \ENDWHILE
    \end{algorithmic}
\end{algorithm}

\begin{table*}[]
\centering
\caption{The descriptions of data sets.}
\label{Datasets}
\begin{tabular}{cccccccccccc}
\hline \hline
Datasets      & Domain & $Lcard$ & \#Instances & \#Features & \#Labels                 & Datasets   & Domain & $Lcard$ & \#Instances & \#Features & \#Labels \\ \hline
health        & text   & 1.66    & 5000        & 612        & \multicolumn{1}{c|}{32}  & bookmarks  & text   & 2.03    & 87856       & 2150       & 208      \\
music\_style  & music  & 1.44    & 6839        & 98         & \multicolumn{1}{c|}{10}  & imdb       & text   & 2.00    & 120919      & 1001       & 28       \\
arts          & text   & 1.64    & 5000        & 462        & \multicolumn{1}{c|}{26}  & eurlex\_sm & text   & 2.21    & 19348       & 5000       & 201      \\
enron         & text   & 3.38    & 1702        & 1001       & \multicolumn{1}{c|}{53}  & mediamill  & video  & 4.38    & 43907       & 120        & 101      \\
corel16k001   & image  & 2.86    & 13766       & 500        & \multicolumn{1}{c|}{153} & tmc2007    & text   & 2.22    & 28956       & 500        & 22       \\
lauguagelog   & text   & 1.18    & 1459        & 1004       & \multicolumn{1}{c|}{75}  & recreation & text   & 1.42    & 5000        & 602        & 22       \\
rcv1\_subset5 & text   & 2.61    & 6000        & 944        & \multicolumn{1}{c|}{101} & medical    & text   & 1.25    & 978         & 1449       & 45       \\ \hline \hline
\end{tabular}
\centering
\end{table*}

\section{Experiments}
In this section, we verify the online classification performance of ELM-NCLD on several datasets from different perspectives. First, we perform extensive experiments on stream data with noisy labels to test the relative performance of competing algorithms. Second, by further considering the potential concept drift in an online environment, we perform concept drift experiments based on the simulated stream data with the noisy and changing label distribution (NCLD). Thirdly, the ablation study is conducted to verify the effectiveness of label reconstruction in accurately scoring labels and the unbiased ranking loss in obtaining the correct score sorting between relevant and irrelevant labels. Next, the experiment of parameter sensitivity is performed to search the optimal value of the regularization parameters. Finally, we analyze the model efficiency based on the time complexity of ELM-NCLD.

For a comprehensive evaluation, a total of fourteen benchmark datasets from different domains are used, whose specific properties are presented in Table \ref{Datasets}, including the domain.  Eight OMC baselines are selected for performance comparison, including FALT, SALT in the paper \cite{Joint}, PAML-I, PAML-II in the paper \cite{MOPAL}, OSML-ELM \cite{ELM}, ELM-CDLL \cite{RELM}, MCIC \cite{MCIC} and MW \cite{Imba} with the corresponding parameters set as suggested in the paper or in the codes, if available. For our ELM-NCLD, as discussed in detail in previous ELM-based works \cite{ELM,RELM}, the number of hidden layer nodes is set as a default value $L=20$ to achieve high efficiency, and we set the regularization factor $\alpha=1$ to limit the model complexity to some extent. Of these, MCIC and MW are replay-based methods and the others are regularization-based methods. Within the regularization-based methods, OSML-ELM, ELM-CDLL and ELM-NCLD utilize ELM as the base model. In order to evaluate the classification performance from different perspectives, three evaluation metrics are used to assess the relative performance between ELM-NCLD and baselines, including Hamming Loss, Micro-F1 and Average Precision, abbreviated as HL, F1 and AP, respectively. Of the three metrics, AP is a ranking-based metric \cite{Review}, which focuses on obtaining the correct ordering of label scores between relevant and irrelevant labels, the other two are classification-based, where HL evaluates the average classification error across all instance-label pairs and F1 indicates the geometric mean of precision and recall.

\begin{table*}[]
\caption{The OMC performance comparison with noisy labels.}
\label{Noisy}
\centering
\begin{tabular}{ccccccccccc}
\hline \hline
                                & Metrics & {\scriptsize ELM-NCLD} & {\scriptsize MCIC} & {\scriptsize MW} & {\scriptsize FALT} & {\scriptsize SALT} & {\scriptsize PAML-I} & {\scriptsize PAML-II} & {\scriptsize OSML-ELM} & {\scriptsize ELM-CDLL} \\ \hline
\multirow{3}{*}{health} & HL      & \textbf{0.0595}                         & 0.2989                              & 0.3683                            & 0.2542                              & 0.4324                              & 0.4736                                & 0.3789                                 & 0.3294                                  & \underline{0.0982}     \\
                                & F1      & \textbf{0.4176}                         & 0.1707                              & 0.1138                            & 0.1322                              & 0.1373                              & 0.1225                                & 0.1282                                 & 0.1672                                  & \underline{0.2296}     \\
                                & AP      & \textbf{0.5819}                         & \underline{0.4736} & 0.1646                            & 0.2190                              & 0.2808                              & 0.2021                                & 0.2082                                 & 0.3786                                  & 0.3827                                  \\ \hline
\multirow{3}{*}{music\_style}   & HL      & \textbf{0.1246}                         & 0.3847                              & 0.3980                            & 0.4483                              & 0.4508                              & 0.1683                                & 0.1683                                 & 0.3037                                  & \underline{0.127}      \\
                                & F1      & \textbf{0.4896}                         & 0.2930                              & 0.2192                            & 0.1873                              & 0.2917                              & 0.0735                                & 0.0735                                 & 0.3236                                  & \underline{0.4856}     \\
                                & AP      & \textbf{0.6090}                         & \underline{0.6015} & 0.3607                            & 0.3337                              & 0.5135                              & 0.3435                                & 0.3344                                 & 0.5425                                  & 0.6013                                  \\ \hline
\multirow{3}{*}{arts}           & HL      & \textbf{0.1032}                         & 0.3024                              & 0.3747                            & 0.3403                              & 0.4387                              & 0.4827                                & 0.4039                                 & 0.3698                                  & \underline{0.1174}     \\
                                & F1      & \textbf{0.2612}                         & \underline{0.2025} & 0.1177                            & 0.1363                              & 0.1733                              & 0.1341                                & 0.1377                                 & 0.1758                                  & 0.1867                                  \\
                                & AP      & \underline{0.3944}     & 0.3511                              & 0.1962                            & 0.2547                              & \textbf{0.4222}                     & 0.2320                                & 0.2386                                 & 0.3247                                  & 0.3276                                  \\ \hline
\multirow{3}{*}{enron}          & HL      & \textbf{0.0748}                         & 0.2820                              & 0.3290                            & 0.3970                              & 0.4068                              & 0.4777                                & 0.3754                                 & \underline{0.1904}     & 0.2003                                  \\
                                & F1      & \textbf{0.4749}                         & 0.2355                              & 0.1676                            & 0.1583                              & 0.2015                              & 0.1483                                & 0.1557                                 & 0.2636                                  & \underline{0.2658}     \\
                                & AP      & \underline{0.4587}     & \textbf{0.4620}                     & 0.2155                            & 0.2335                              & 0.4127                              & 0.2280                                & 0.2264                                 & 0.3524                                  & 0.3652                                  \\ \hline
\multirow{3}{*}{core15k}        & HL      & \textbf{0.0255}                         & 0.3007                              & 0.3417                            & 0.3979                              & 0.4854                              & 0.4901                                & 0.3024                                 & \underline{0.2565}     & 0.4036                                  \\
                                & F1      & \textbf{0.1074}                         & \underline{0.0265} & 0.0236                            & 0.0233                              & 0.0226                              & 0.0197                                & 0.0203                                 & 0.0247                                  & 0.0234                                  \\
                                & AP      & \textbf{0.1247}                         & \underline{0.0602} & 0.0451                            & 0.0305                              & 0.0360                              & 0.0278                                & 0.0278                                 & 0.0408                                  & 0.0411                                  \\ \hline
\multirow{3}{*}{corel16k001}    & HL      & \textbf{0.0653}                         & 0.3007                              & 0.3543                            & 0.3979                              & 0.4742                              & 0.4902                                & 0.3528                                 & 0.3207                                  & \underline{0.2895}     \\
                                & F1      & \textbf{0.1493}                         & \underline{0.0582} & 0.0467                            & 0.0443                              & 0.0466                              & 0.0387                                & 0.0393                                 & 0.0558                                  & 0.0581                                  \\
                                & AP      & \textbf{0.1960}                         & \underline{0.1534} & 0.1029                            & 0.0912                              & 0.1249                              & 0.0704                                & 0.0704                                 & 0.1273                                  & 0.1280                                  \\ \hline
\multirow{3}{*}{languagelog}    & HL      & \textbf{0.0943}                         & 0.2873                              & 0.3733                            & 0.3426                              & 0.4817                              & 0.4614                                & 0.3267                                 & 0.3701                                  & \underline{0.2328}     \\
                                & F1      & \textbf{0.0612}                         & 0.0290                              & 0.0333                            & 0.0391                              & \underline{0.0400} & 0.0388                                & 0.0390                                 & 0.0329                                  & 0.0319                                  \\
                                & AP      & \underline{0.0940}     & 0.0534                              & 0.0906                            & \textbf{0.1223}                     & 0.0604                              & 0.0849                                & 0.0848                                 & 0.0715                                  & 0.0671                                  \\ \hline
\multirow{3}{*}{rcv1\_subset5}  & HL      & \textbf{0.0373}                         & 0.3203                              & 0.3560                            & 0.3410                              & 0.4122                              & 0.4855                                & 0.3505                                 & \underline{0.0525}     & 0.2452                                  \\
                                & F1      & \textbf{0.1920}                         & 0.0665                              & 0.0746                            & 0.0676                              & 0.0570                              & 0.0587                                & 0.0630                                 & \underline{0.1156}     & 0.0749                                  \\
                                & AP      & \underline{0.2261}     & 0.1857                              & 0.1597                            & 0.2619                              & \textbf{0.4069}                     & 0.1927                                & 0.1934                                 & 0.1999                                  & 0.2016                                  \\ \hline
\multirow{3}{*}{bookmarks}      & HL      & \underline{0.2695}     & 0.3070                              & 0.3652                            & 0.3977                              & 0.4921                              & 0.4712                                & 0.2977                                 & \textbf{0.2055}                         & 0.3234                                  \\
                                & F1      & \underline{0.0255}     & 0.0211                              & 0.0219                            & 0.0232                              & 0.0229                              & 0.0229                                & 0.0235                                 & \textbf{0.0267}                         & 0.0251                                  \\
                                & AP      & \textbf{0.1740}                         & 0.0360                              & 0.0696                            & 0.0546                              & 0.0793                              & 0.0449                                & 0.0450                                 & \underline{0.1519}     & 0.1516                                  \\ \hline
\multirow{3}{*}{imdb}           & HL      & \underline{0.0929}     & 0.3022                              & 0.3625                            & 0.3046                              & 0.4656                              & 0.4880                                & 0.4333                                 & 0.4566                                  & \textbf{0.0878}                         \\
                                & F1      & \textbf{0.3025}                         & \underline{0.2309} & 0.1280                            & 0.1289                              & 0.1726                              & 0.1374                                & 0.1368                                 & 0.1819                                  & 0.1970                                  \\
                                & AP      & \textbf{0.4530}                         & \underline{0.3710} & 0.1878                            & 0.1968                              & 0.2680                              & 0.1950                                & 0.1944                                 & 0.3657                                  & 0.3658                                  \\ \hline
\multirow{3}{*}{eurlex\_sm}     & HL      & \underline{0.2645}     & 0.2995                              & 0.3607                            & 0.3666                              & 0.4721                              & 0.4683                                & 0.2957                                 & \textbf{0.2584}                         & 0.3176                                  \\
                                & F1      & \textbf{0.0499}                         & 0.0322                              & 0.0318                            & 0.0400                              & 0.0341                              & 0.0319                                & 0.0378                                 & \underline{0.0422}     & 0.0387                                  \\
                                & AP      & \textbf{0.2110}                         & 0.0851                              & 0.0706                            & \underline{0.1769} & 0.1422                              & 0.1152                                & 0.1156                                 & 0.1634                                  & 0.1632                                  \\ \hline
\multirow{3}{*}{mediamill}      & HL      & \textbf{0.0351}                         & 0.2902                              & 0.3388                            & 0.4206                              & 0.3956                              & 0.2775                                & 0.2710                                 & \underline{0.0575}     & 0.1858                                  \\
                                & F1      & \textbf{0.5309}                         & 0.1713                              & 0.1181                            & 0.1182                              & 0.1415                              & 0.1519                                & 0.1546                                 & \underline{0.4656}     & 0.2442                                  \\
                                & AP      & \textbf{0.6167}                         & 0.5747                              & 0.2103                            & 0.4154                              & 0.4835                              & 0.2680                                & 0.2584                                 & \underline{0.6100}     & 0.6030                                  \\ \hline
\multirow{3}{*}{mirflickr}      & HL      & \textbf{0.1491}                         & 0.3417                              & 0.3962                            & 0.4895                              & 0.4819                              & 0.4916                                & 0.4916                                 & \underline{0.1728}     & \underline{0.1728}     \\
                                & F1      & \textbf{0.7216}                         & 0.5030                              & 0.3698                            & 0.3664                              & 0.3609                              & 0.3911                                & 0.3889                                 & 0.6266                                  & \underline{0.6393}     \\
                                & AP      & \textbf{0.8424}                         & 0.7070                              & 0.5076                            & 0.4214                              & 0.4690                              & 0.4571                                & 0.4299                                 & 0.7844                                  & \underline{0.7870}     \\ \hline
\multirow{3}{*}{tmc2007}        & HL      & \underline{0.1038}     & 0.3028                              & 0.3665                            & 0.2819                              & 0.3835                              & 0.4572                                & 0.3738                                 & 0.1920                                  & \textbf{0.0978}                         \\
                                & F1      & \textbf{0.5132}                         & 0.3104                              & 0.2199                            & 0.2635                              & 0.2861                              & 0.2374                                & 0.2489                                 & 0.3961                                  & \underline{0.4929}     \\
                                & AP      & \textbf{0.6204}                         & 0.5309                              & 0.2993                            & 0.4623                              & 0.5087                              & 0.4635                                & 0.4620                                 & \underline{0.5759}     & 0.5765                                  \\ \hline
\multirow{3}{*}{recreation}         & HL      & \textbf{0.1222}                         & 0.3126                              & 0.3712                            & 0.2680                              & 0.4672                              & 0.4712                                & 0.4108                                 & 0.2220                                  & \underline{0.1330}     \\
                                & F1      & \textbf{0.2148}                         & 0.1837                              & 0.1278                            & 0.1527                              & 0.1730                              & 0.1462                                & 0.1497                                 & \underline{0.1894}     & 0.1716                                  \\
                                & AP      & 0.3404                                  & 0.3189                              & 0.2175                            & 0.3877                              & \textbf{0.4232}                     & 0.3866                                & \underline{0.3889}    & 0.3155                                  & 0.3160                                  \\ \hline
\multirow{3}{*}{medical}        & HL      & \textbf{0.0279}                         & 0.3400                              & 0.3615                            & 0.2657                              & 0.4278                              & 0.4795                                & 0.3634                                 & 0.3037                                  & \underline{0.1820}     \\
                                & F1      & \textbf{0.2424}                         & 0.0589                              & 0.0958                            & 0.0921                              & 0.0993                              & 0.0743                                & 0.0781                                 & 0.1052                                  & \underline{0.1340}     \\
                                & AP      & \textbf{0.3489}                         & 0.2506                              & 0.2531                            & 0.1715                              & 0.2867                              & 0.1552                                & 0.1529                                 & 0.2962                                  & \underline{0.3059}     \\ \hline \hline
\end{tabular}
\centering
\end{table*}

\begin{table}[!t]
\tabcolsep 0.04in 
\centering
\setlength{\belowcaptionskip}{0.5cm} 
\caption{Summary of the Friedman statistics $F_F$ between ELM-NCLD and other algorithms in terms of each evaluation criterion and the critical value ($\alpha = 0.1$).}
\label{F}
\begin{tabular}{@{}ccccc@{}}
\toprule \toprule
Evaluation Criteria & \#Tests  &\#Algorithms  & $F_F$       & Critical value \\ \toprule
Fig. \ref{HL}: HL    &16 &9    & 40.0603  & 2.6400 \\
Fig. \ref{F1}: F1 &16 &9 & 23.9881  & 2.6400 \\
Fig. \ref{AP}: AP    &16 &9    & 13.6781      & 2.6400\\
Fig. \ref{HL_r}: HL(NCLD)    &12 &11    &  81.7796         & 2.8600 \\
Fig. \ref{F1_r}: F1(NCLD)     &12 &11     &  25.0298   & 2.8600    \\ 
Fig. \ref{AP_r}: AP(NCLD)  &12 &11 & 8.5996 & 2.8600  \\ 
\bottomrule \bottomrule
\end{tabular}
\end{table}

\newcommand{\w}{5cm}
\newcommand{\h}{2.5cm}
\begin{figure*}[htbp]
  \centering
  \subfloat[HL]{\includegraphics[width=\w,height=\h]{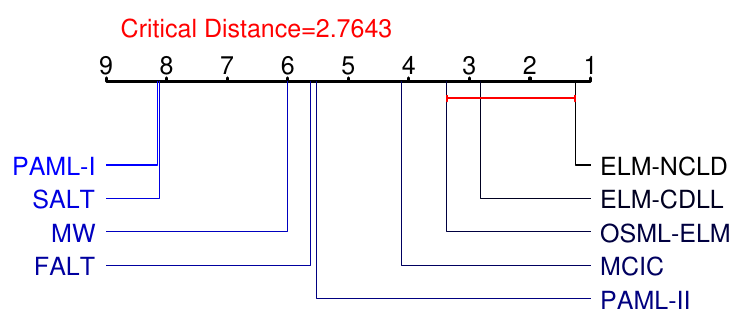} \label{HL}}
  \subfloat[F1]{\includegraphics[width=\w,height=\h]{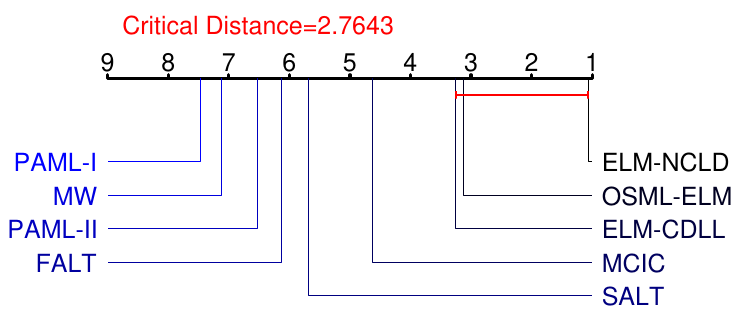} \label{F1}} \quad
  \subfloat[AP]{\includegraphics[width=\w,height=\h]{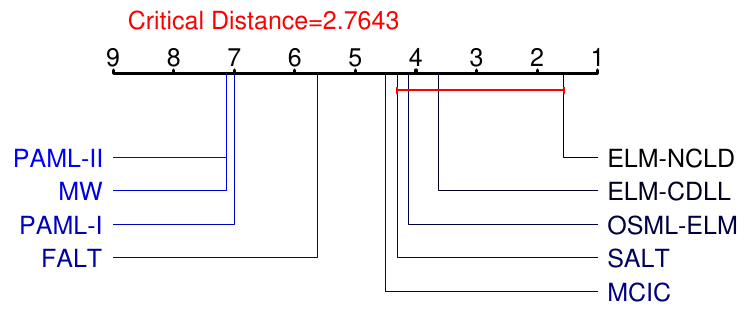} \label{AP}} \quad
  \subfloat[HL(NCLD)]{\includegraphics[width=\w,height=\h]{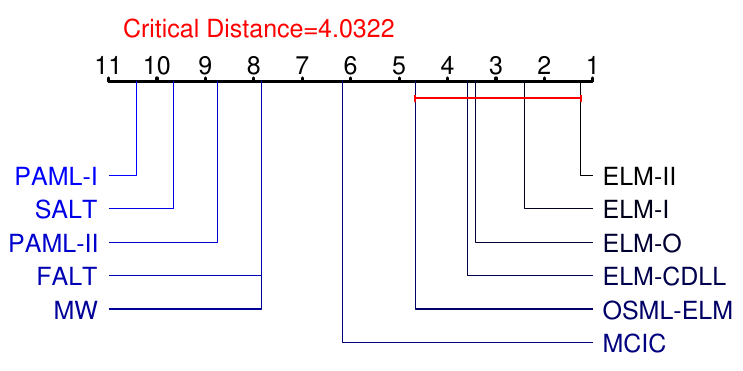}\label{HL_r}} \quad
  \subfloat[F1(NCLD)]{\includegraphics[width=\w,height=\h]{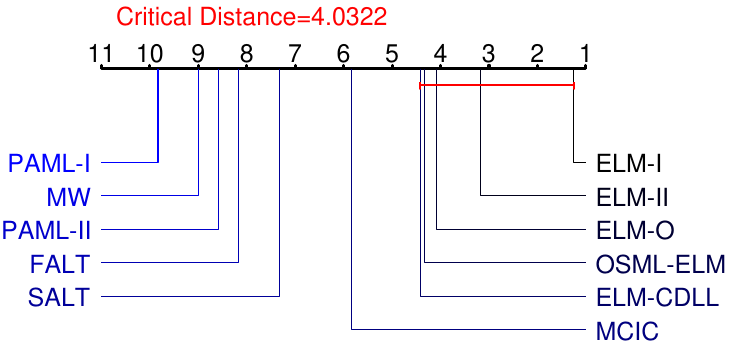} \label{F1_r}}
  \subfloat[AP(NCLD)]{\includegraphics[width=\w,height=\h]{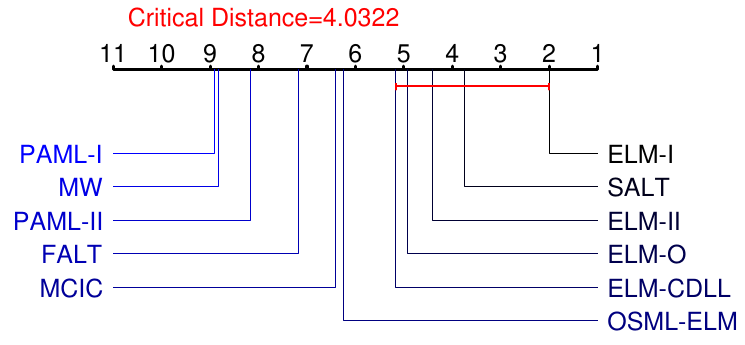} \label{AP_r}}
  \caption{Comparison of ELM-NCLD against other algorithms with Nemenyi test.}
  \label{CD}
\end{figure*}

\subsection{The Effectiveness with Noisy Labels}
Without considering the distribution changes in labels, we first evaluate the online classification performance regarding general noisy data stream in this section. To simulate the noisy data stream, we first inject noisy labels for each class label by randomly flipping the ground truth relevant and irrelevant labels over the noisy rate $\rho^j_{+}$ and $\rho^j_{-}, j \in [q]$, which is randomly chosen within the range $[0.2,0.4]$. We then randomly divide the batch data into different data chunks of size $N = 500$ and process these data chunks sequentially to perform online classification. The specific experimental results are presented in Table \ref{Noisy}, where the bold and underlined values indicate the best and the second best result respectively. In addition, the Friedman test and the Nemenyi test \cite{Friedman-Test} are performed to give the relative performance among these comparison algorithms with respect to HL, F1 and AP, whose results are shown in Table \ref{Noisy} and Fig. \ref{HL}-\ref{AP}:   

From these results, we have the following discussions: 

1) \textbf{ELM-NCLD} \emph{against} \textbf{Other} methods: From Figs. \ref{HL}-\ref{AP} we can see that ELM-NCLD ranks first on all three evaluation metrics in the significance test and significantly outperforms other methods except OSML-ELM and ELM-CDLL, and it ranks \emph{1st} in 75.0\% of cases and \emph{2nd} in 18.8\% of cases on three metrics from Table \ref{Noisy}. The reason for this is that other competing algorithms are mainly designed to solve the fully supervised task, whose performance is severely affected by the potentially noisy labels in the stream data, while ELM-NCLD establishes robustness to noisy labels during the sequential process of classification and model update using the noisy data. Based on the label scoring and ranking framework, on the one hand, the label scores of each instance are reconstructed by its observed labels and the label scores of neighbouring instances, which indirectly filters out the incredible labels and is beneficial for model fitting. On the other hand, to make it reasonable to predict relevant labels with the fixed threshold of 0, the unbiased ranking loss term is derived and applied to be integrated into the whole optimization framework to achieve a reliable ranking order given noisy observations. 

2) \textbf{ELM-based} \emph{against} \textbf{other} methods: For all three evaluation metrics, the ELM-based methods are always in the top three, and the performance of OSML-ELM and ELM-CDLL is comparable to that of ELM-CDLL. Although OSML-ELM and ELM-CDLL do not design customized mechanisms to handle online noisy labels like other methods without using ELM, they significantly outperform others due to the intrinsic superiority of ELM. Firstly, ELM uses non-linear feature mapping to reduce the high dimensionality of online multi-label data, which is more general and universal to real-world data compared to linear methods such as FALT, PAML-I. Secondly, the dimensionality reduction of the online data not only speeds up the model update, but also improves the online classification accuracy by removing irrelevant and redundant features.

3) \textbf{Regularization-based} \emph{against} \textbf{Replay-based} methods: A total of two replay-based methods are used for comparison, including MCIC and MW. For MW, a local data pool is built and updated with fixed positive and negative instance ratios for each label. However, the online noisy labels will introduce false positive and negative instances into the data pool, which intuitively will degrade the classification performance and thus achieve a lower performance ranking. For MCIC from Fig. \ref{CD}, it just ranks after ELM-based methods and outperforms other regularisation-based methods in terms of HL and F1, and only loses to a regularisation-based method (SALT) on AP. The possible reason for this is that MCIC computes and stores the mature clusters as the data summary, rather than each instance in the original feature space as MW does, so when classifying based on the data summary, the model is less influenced by the noisy instances that are integrated to form the cluster point. 

\begin{table*}[]
\caption{The OMC performance comparison under NCLD.}
\label{NCLD}
\centering
\begin{tabular}{ccccccccccccc}
\hline \hline
Data sets                         & Metrics & {\scriptsize ELM-O} & {\scriptsize ELM-I} & {\scriptsize ELM-II} & {\scriptsize MCIC} & {\scriptsize MW}   & {\scriptsize FALT} & {\scriptsize SALT} & {\scriptsize PAML-I} & {\scriptsize PAML-II} & {\scriptsize OSML-ELM} & {\scriptsize ELM-CDLL} \\ \hline
\multirow{3}{*}{mediamill\_a}     & HL      & 0.0171                                  & \textbf{0.0147}                      & \underline{0.0163}   & 0.2821                              & 0.3564                              & 0.4546                              & 0.4777                              & 0.3440                                & 0.3416                                 & 0.2207                                  & 0.2043                                  \\
                                  & F1      & \underline{0.4234}     & \textbf{0.5566}                      & 0.4197                                & 0.0796                              & 0.0420                              & 0.0258                              & 0.0464                              & 0.0328                                & 0.0330                                 & 0.0968                                  & 0.1021                                  \\
                                  & AP      & 0.5227                                  & \textbf{0.6356}                      & 0.5358                                & 0.4788                              & 0.1146                              & 0.6169                              & \underline{0.6299} & 0.5209                                & 0.5328                                 & 0.5018                                  & 0.5020                                  \\ \hline
\multirow{3}{*}{mediamill\_d}     & HL      & 0.0175                                  & \textbf{0.0140}                      & \underline{0.0162}   & 0.2901                              & 0.3611                              & 0.4536                              & 0.4689                              & 0.2966                                & 0.2681                                 & 0.0427                                  & 0.1979                                  \\
                                  & F1      & 0.5232                                  & \textbf{0.5786}                      & \underline{0.5368}   & 0.0838                              & 0.0396                              & 0.0474                              & 0.0573                              & 0.0492                                & 0.0534                                 & 0.3435                                  & 0.1166                                  \\
                                  & AP      & 0.6012                                  & \textbf{0.6586}                      & \underline{0.6172}   & 0.5663                              & 0.1120                              & 0.0946                              & 0.2151                              & 0.0788                                & 0.0744                                 & 0.5936                                  & 0.6057                                  \\ \hline
\multirow{3}{*}{mirflickr\_a}     & HL      & \underline{0.1482}     & 0.1521                               & \textbf{0.1443}                       & 0.3420                              & 0.3860                              & 0.4251                              & 0.4203                              & 0.4961                                & 0.4924                                 & 0.2241                                  & 0.1714                                  \\
                                  & F1      & 0.7055                                  & \textbf{0.7263}                      & \underline{0.7132}   & 0.4891                              & 0.3619                              & 0.4137                              & 0.4307                              & 0.3895                                & 0.3949                                 & 0.6156                                  & 0.6232                                  \\
                                  & AP      & 0.8208                                  & \textbf{0.8298}                      & \underline{0.8276}   & 0.6733                              & 0.5004                              & 0.4347                              & 0.6476                              & 0.5347                                & 0.5405                                 & 0.7612                                  & 0.7624                                  \\ \hline
\multirow{3}{*}{mirflickr\_d}     & HL      & 0.1827                                  & \underline{0.1598}  & \textbf{0.1549}                       & 0.2722                              & 0.3908                              & 0.4293                              & 0.4039                              & 0.4797                                & 0.4774                                 & 0.1894                                  & 0.1623                                  \\
                                  & F1      & 0.6751                                  & \textbf{0.7159}                      & \underline{0.7110}   & 0.5884                              & 0.3485                              & 0.4060                              & 0.4576                              & 0.3815                                & 0.3826                                 & 0.6522                                  & 0.6692                                  \\
                                  & AP      & 0.7779                                  & \textbf{0.8207}                      & \underline{0.8018}   & 0.7247                              & 0.4857                              & 0.6395                              & 0.5765                              & 0.6109                                & 0.6383                                 & 0.7642                                  & 0.7656                                  \\ \hline
\multirow{3}{*}{rcv1\_subset5\_a} & HL      & \underline{0.0367}     & 0.0397                               & \textbf{0.0313}                       & 0.3155                              & 0.3487                              & 0.3252                              & 0.4393                              & 0.4897                                & 0.3569                                 & 0.0767                                  & 0.2574                                  \\
                                  & F1      & 0.1261                                  & \textbf{0.1416}                      & \underline{0.1407}   & 0.0430                              & 0.0794                              & 0.0536                              & 0.0359                              & 0.0511                                & 0.0548                                 & 0.0561                                  & 0.0472                                  \\
                                  & AP      & 0.1956                                  & 0.2349                               & 0.1946                                & 0.1913                              & \underline{0.2452} & 0.2286                              & \textbf{0.5640}                     & 0.1654                                & 0.1655                                 & 0.1448                                  & 0.1502                                  \\ \hline
\multirow{3}{*}{rcv1\_subset5\_d} & HL      & 0.0506                                  & \underline{0.0426}  & \textbf{0.0368}                       & 0.3169                              & 0.3496                              & 0.3421                              & 0.4043                              & 0.4919                                & 0.3538                                 & 0.0690                                  & 0.2575                                  \\
                                  & F1      & \underline{0.1282}     & \textbf{0.1405}                      & 0.1056                                & 0.0505                              & 0.0674                              & 0.0460                              & 0.0376                              & 0.0406                                & 0.0419                                 & 0.0935                                  & 0.0565                                  \\
                                  & AP      & 0.2876                                  & 0.2860                               & 0.2767                                & 0.1164                              & 0.2536                              & \underline{0.3674} & \textbf{0.4733}                     & 0.2407                                & 0.2413                                 & 0.1421                                  & 0.1431                                  \\ \hline
\multirow{3}{*}{recreation\_a}    & HL      & 0.0885                                  & \underline{0.0866}  & \textbf{0.0834}                       & 0.3052                              & 0.3713                              & 0.3219                              & 0.4605                              & 0.4690                                & 0.4072                                 & 0.4527                                  & 0.1214                                  \\
                                  & F1      & 0.1472                                  & \underline{0.1654}  & 0.1391                                & \textbf{0.1895}                     & 0.1231                              & 0.1526                              & 0.1619                              & 0.1391                                & 0.1445                                 & 0.1554                                  & 0.1765                                  \\
                                  & AP      & 0.2952                                  & 0.3075                               & 0.2994                                & \underline{0.3496} & 0.2219                              & 0.2635                              & \textbf{0.4430}                     & 0.2918                                & 0.2912                                 & 0.3263                                  & 0.3287                                  \\ \hline
\multirow{3}{*}{recreation\_d}    & HL      & 0.2529                                  & 0.1378                               & \textbf{0.0934}                       & 0.3177                              & 0.3682                              & 0.3298                              & 0.5065                              & 0.4799                                & 0.4213                                 & 0.2227                                  & \underline{0.1294}     \\
                                  & F1      & 0.0973                                  & \textbf{0.2207}                      & 0.1339                                & 0.1644                              & 0.1090                              & 0.1311                              & 0.1397                              & 0.1238                                & 0.1255                                 & 0.1701                                  & \underline{0.1731}     \\
                                  & AP      & 0.1605                                  & \underline{0.3214}  & 0.2209                                & 0.2920                              & 0.2165                              & 0.2315                              & \textbf{0.3582}                     & 0.2315                                & 0.2333                                 & 0.3008                                  & 0.3015                                  \\ \hline
\multirow{3}{*}{arts\_a}          & HL      & 0.1252                                  & 0.1207                               & \textbf{0.1074}                       & 0.3054                              & 0.3696                              & 0.2894                              & 0.3889                              & 0.4726                                & 0.3978                                 & 0.2037                                  & \underline{0.1118}     \\
                                  & F1      & \underline{0.2602}     & \textbf{0.2871}                      & 0.2544                                & 0.1554                              & 0.1141                              & 0.1491                              & 0.1489                              & 0.1321                                & 0.1370                                 & 0.1613                                  & 0.1724                                  \\
                                  & AP      & 0.3940                                  & \underline{0.4298}  & 0.3935                                & 0.3557                              & 0.2221                              & 0.2993                              & \textbf{0.4584}                     & 0.2984                                & 0.3149                                 & 0.3178                                  & 0.3215                                  \\ \hline
\multirow{3}{*}{arts\_d}          & HL      & 0.1441                                  & \underline{0.0991}  & \textbf{0.0784}                       & 0.3070                              & 0.3683                              & 0.3179                              & 0.4512                              & 0.4871                                & 0.4162                                 & 0.1914                                  & 0.1185                                  \\
                                  & F1      & 0.1226                                  & \textbf{0.2960}                      & \underline{0.1657}   & 0.1389                              & 0.1092                              & 0.1142                              & 0.1305                              & 0.1106                                & 0.1126                                 & 0.1629                                  & 0.1549                                  \\
                                  & AP      & 0.2453                                  & \textbf{0.3985}                      & 0.2435                                & 0.2589                              & 0.2197                              & 0.2628                              & \underline{0.3465} & 0.2376                                & 0.2386                                 & 0.2654                                  & 0.2666                                  \\ \hline
\multirow{3}{*}{tmc2007\_a}       & HL      & 0.1005                                  & 0.1056                               & \textbf{0.0980}                       & 0.3281                              & 0.3525                              & 0.3034                              & 0.4351                              & 0.4702                                & 0.3859                                 & 0.1330                                  & \underline{0.0999}     \\
                                  & F1      & \underline{0.4705}     & \textbf{0.5210}                      & 0.4560                                & 0.2319                              & 0.2250                              & 0.1969                              & 0.1924                              & 0.1857                                & 0.1902                                 & 0.3991                                  & 0.4177                                  \\
                                  & AP      & 0.5934                                  & \textbf{0.6276}                      & \underline{0.5966}   & 0.4642                              & 0.3341                              & 0.3215                              & 0.2995                              & 0.3251                                & 0.3272                                 & 0.5211                                  & 0.5222                                  \\ \hline
\multirow{3}{*}{tmc2007\_d}       & HL      & 0.1132                                  & \underline{0.1041}  & \textbf{0.0975}                       & 0.3236                              & 0.3511                              & 0.4128                              & 0.3566                              & 0.4697                                & 0.4058                                 & 0.0860                                  & 0.1104                                  \\
                                  & F1      & 0.5115                                  & \underline{0.5281}  & \textbf{0.5411}                       & 0.2654                              & 0.2203                              & 0.1991                              & 0.2985                              & 0.1970                                & 0.2009                                 & 0.5265                                  & 0.5036                                  \\
                                  & AP      & 0.6263                                  & \underline{0.6462}  & 0.6412                                & 0.5349                              & 0.3384                              & 0.3023                              & \textbf{0.6685}                     & 0.2922                                & 0.2901                                 & 0.6133                                  & 0.6142                                  \\ \hline \hline
\end{tabular}
\centering
\end{table*}

\subsection{The Effectiveness under NCLD}
In this section, we compare ELM-OMLL with baselines with respect to the classification performance under noisy and changing label distribution (NCLD), where two special cases are considered namely concept growth and concept reduction. The former indicates single-label instances first, then multi-label, while in the latter case the opposite is true. To additionally simulate the label noise, we then inject the noisy labels for these concept drift data sets according to the noise rates defined in the previous section. Here, six data sets are utilized to simulate the two NCLD cases including mediamill, mirflickr, rcv1\_subset5, recreation, arts and tmc2007 as the representative, since the similar results can be observed on other data sets. We denote the synthetic data sets with concept growth or concept reduction as the \emph{name} with the suffix `\_a' or `\_b' respectively. The notation ``ELM-O" is used to denote the original version of our method without the concept drift adaption, and the versions leveraging the \emph{retrain} and \emph{adjustment} adaption strategies are denoted as ``ELM-I" and ``ELM-II" respectively. Table \ref{NCLD} presents that specific comparison results between the three variants of ELM-NCLD and baselines, and Figs. \ref{HL_r}-\ref{AP_r} give the corresponding significance test results, from which we have the following analysis:

1) \textbf{ELM-NCLD} Variants \emph{against} \textbf{Other} methods: Considering ELM-O, ELM-I and ELM-II together in relation to the Table \ref{NCLD}, they rank \emph{1st}, \emph{2nd} in 80.6\%, 8.3\% of the total cases, the reason for this can be attributed to two aspects: i) Compared to other methods, ELM-NCLD can deal with the potential noisy labels, thus achieving the relatively high classification performance; ii) In terms of ELM-I and ELM-II, they detect and adapt to the ground-truth concept drift from the noisy label distribution, thus establishing the robustness to noisy and changing label distribution. From Figs. \ref{HL_r}-\ref{AP_r}, three ELM-NCLD variants rank in the top three on HL and F1, the only exception being that SALT outperforms ELM-O and ELM-II, but loses to ELM-I on AP. The possible reason for this is that SALT uses the Hessian to adaptively determine the learning rate, thus achieving a fast convergence speed for the distribution after drift. 

2) \textbf{ELM-O} \emph{against} \textbf{ELM-I} \& \textbf{ELM-II}: Based on the self-comparison among ELM-NCLD variants from Figs. \ref{HL_r}-\ref{AP_r}, ELM-I and ELM-II always outperform ELM-O verifying the following two facts: i) The idea of utilizing the unbiased estimation of ground-truth label cardinality is valid on detecting the NCLD with a proper confidence value $\delta$ for the Hoeffding inequality; ii) Both two concept drift adaption strategies are beneficial to quickly adapt to the new distribution including \emph{retrain} and \emph{adjustment}, corresponding to ELM-I and ELM-II respectively. In addition, it is worth noting that ELM-I is superior to ELM-II on HL but inferior to ELM-II on F1 and AP. The only difference between ELM-I and ELM-II is that ELM-I leverages the old label scoring from the last distribution, which leads to advantages on different metrics. Specifically, the label scoring information without considering the label ranking is \emph{consistent} with respect to HL \cite{Cons}, which is why ELM-I surpasses ELM-II on HL, and the reason why ELM-II wins on AP and F1 can be reasonably attributed to the fact that the ranking relationship intrinsically hidden in the old label scoring information interferes with modelling the new label ranking information. 

\begin{figure*}
\centering
\subfloat[languagelog]{\includegraphics[width=0.25\textwidth]{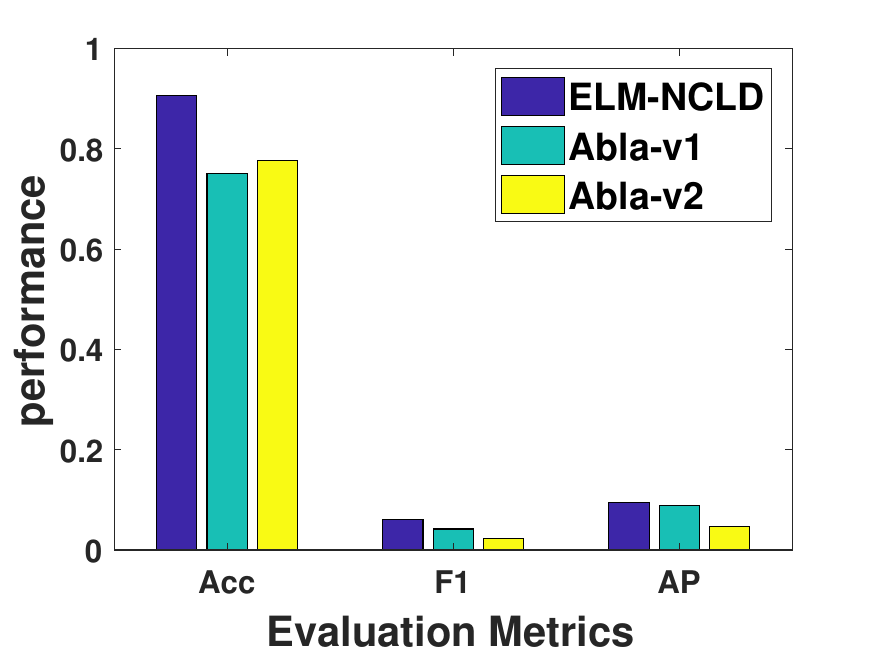}}
\subfloat[arts]{\includegraphics[width=0.25\textwidth]{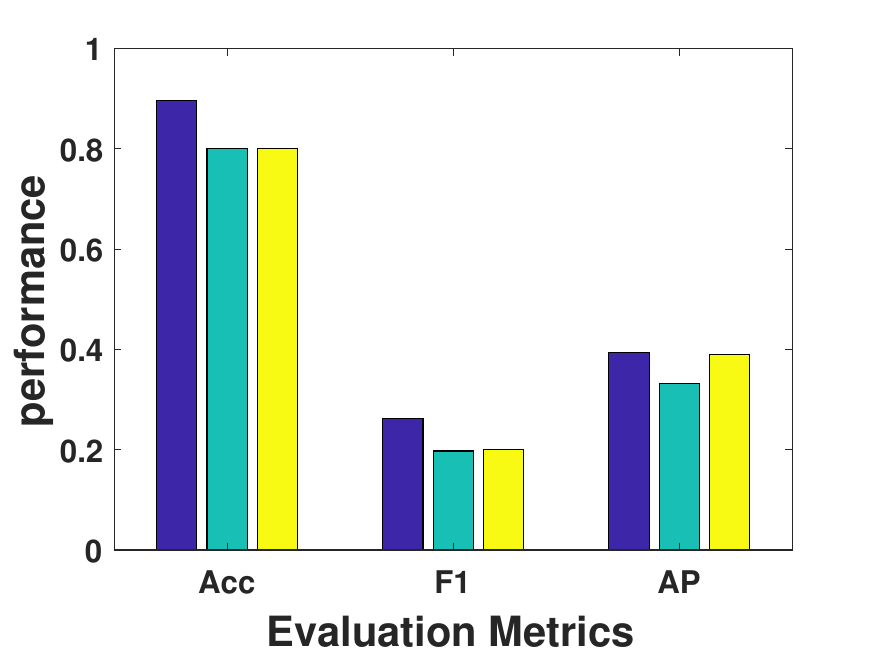}}
\subfloat[medical]{\includegraphics[width=0.25\textwidth]{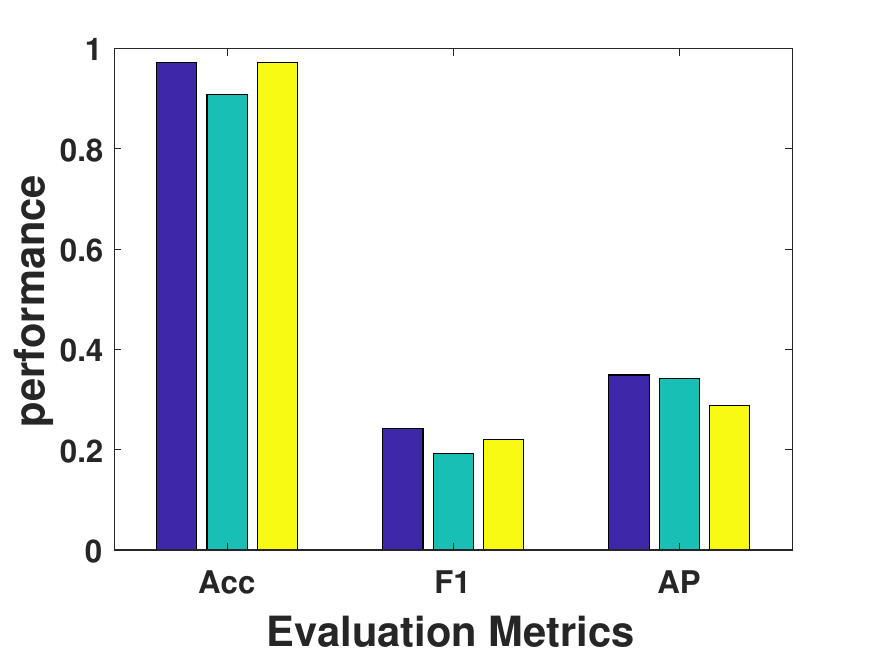}}
\subfloat[corel16k001]{\includegraphics[width=0.25\textwidth]{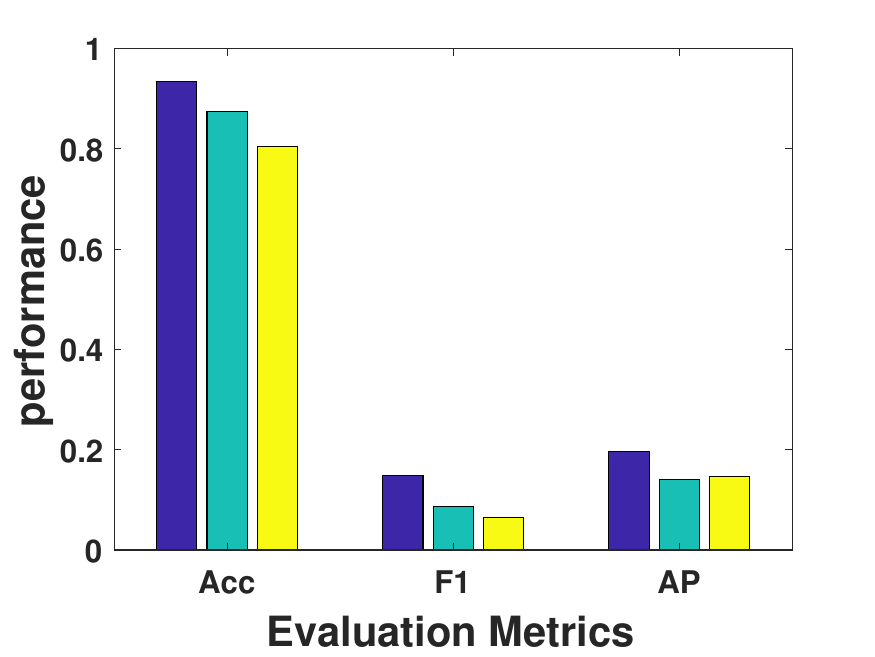}} \quad
\caption{Ablation performance comparison.}
\label{Abla}
\end{figure*} 

\begin{figure*}
\centering
\subfloat[rcv1\_subset5]{\includegraphics[width=0.45\textwidth]{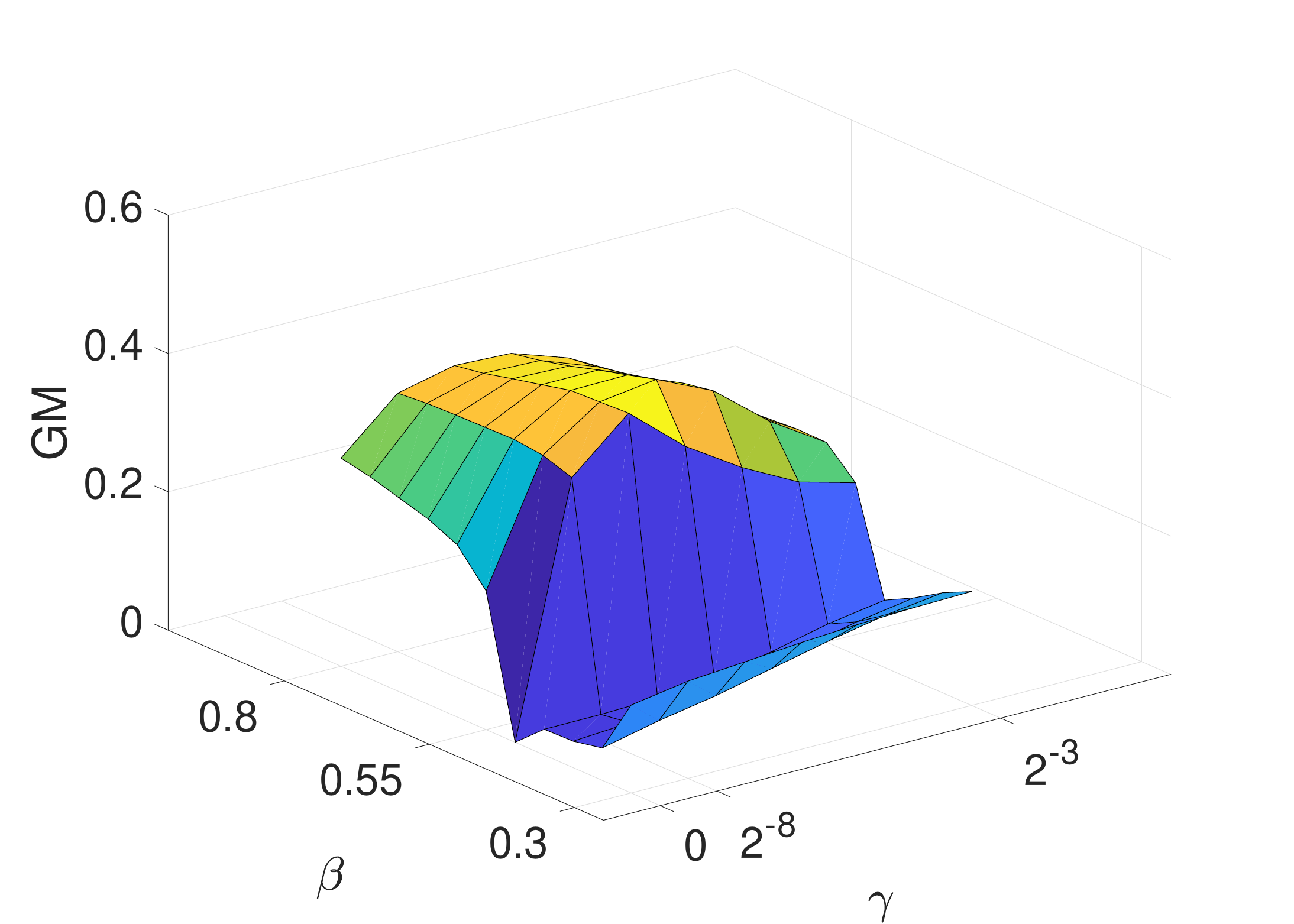}\label{Sens1}}
\subfloat[medical]{\includegraphics[width=0.45\textwidth]{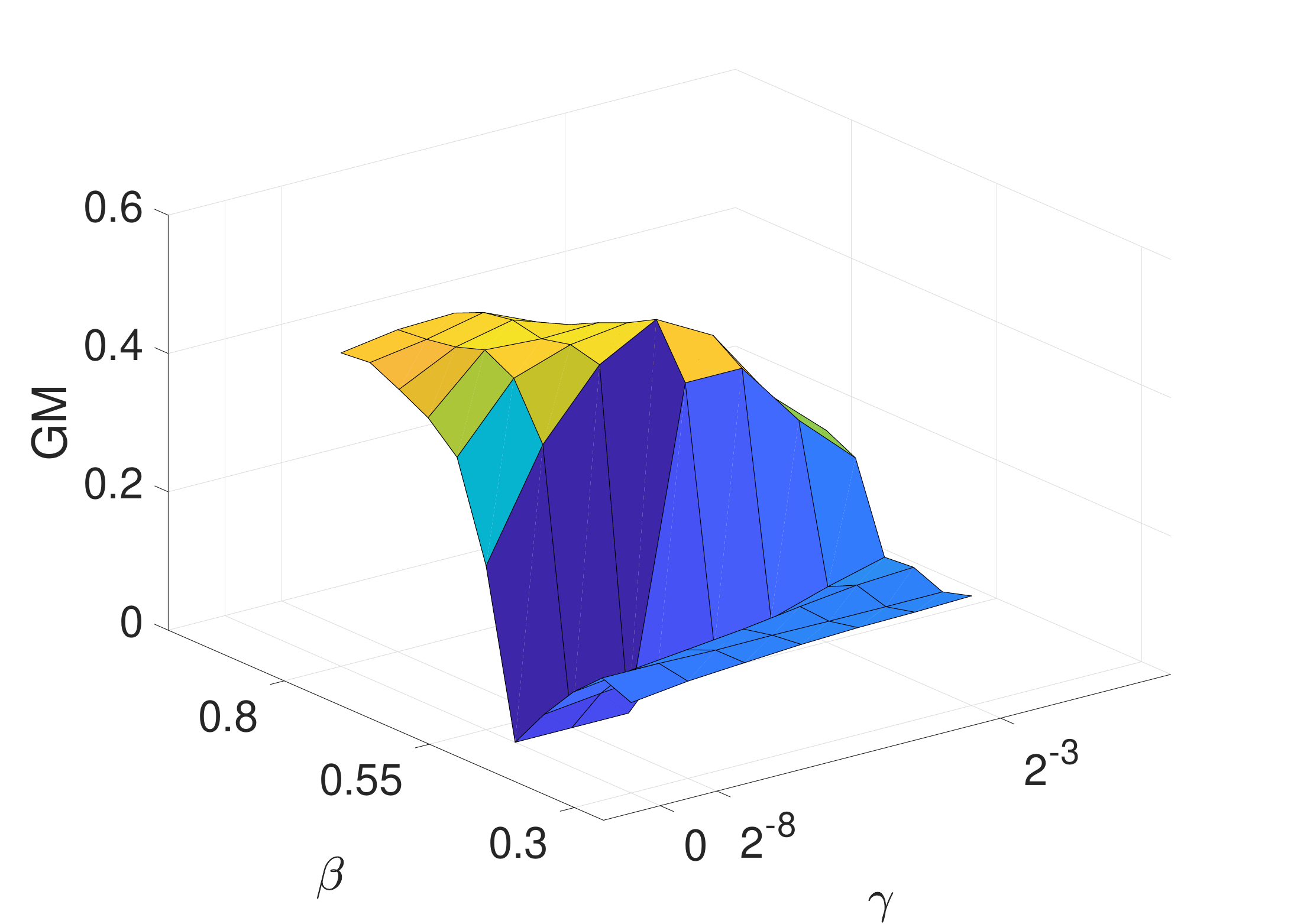}\label{Sens2}}
\quad
\caption{Parameter sensitivity analysis.}
\label{Sens}
\end{figure*}
\subsection{Ablation Study}
In this section, we verify the effectiveness of certain components of our objective. Specifically, we perform ablation experiments on to eliminate or tune certain loss terms and compare the performance with the original version. Two ablation versions are formalized: 1) Abla-v1: The objective corresponding to the Eq. \eqref{N3} with $\beta=1$, in other words, this version does not reconstruct the label scores utilizing the graph built on the feature space, but only regards the observed labels as the the label scores. 2) Abla-v2: Without considering the wrong label ranking relationship led by potential noisy labels, the ranking loss term with respect to Eq. \eqref{RL} rather than Eq. \eqref{RL_New} is integrated into the final objective in Eq. \eqref{N3}. Four data sets are used for comparison since similar results are achieved on other data sets, including languagelog, arts, medical and corel16k001. We present Acc = 1 - HL for clarity, the experimental results are shown in Fig. \ref{Abla}.

From Fig. \ref{Abla}, we can see that ELM-NCLD outperforms Abla-v1 in all cases, which confirms the effectiveness of reconstructing the label scores using feature graph information, because in the case of noisy labels, it is risky to consider the observed labels of each instance as label scores. Based on the fact that the neighbours of each point in the feature graph are more likely to share the same labels with it, it is beneficial to use the label scores of the nearest neighbours in the feature space as partial support information, as shown in Eq. \eqref{P1}. In addition, ELM-NCLD is also always superior to Abla-v1, indicating that the unbiased estimation of the ranking loss with respect to Eq. \eqref{RL_New} really helps to obtain an accurate ranking order compared to the original term in Eq. \eqref{RL}. This also indirectly verifies that the conditional independence assumption used in the derivation of \eqref{RL_New} basically holds for the robust estimation of ranking loss.

\subsection{Parameter Sensitivity Analysis}
In this section we search for the optimal value of the regularization factors $\beta$ and $\gamma$ using a mesh search method, recalling that $\beta$ controls the weighting of the label reconstruction between the observed labels and the label scores of neighbours in the feature graph, $\gamma$ measures the balancing weight of the label ranking term compared to the label scoring term. The $\beta$ is chosen from the range [0.3,0.8] with a fixed step of 0.05; although the theoretical range of $\beta$ is [0,1], we reduce the search range to [0.3,0.8] for efficiency reasons due to the poor performance with a too large or too small $\beta$. As for $\gamma$, we choose it from the range $\{0,2^{-8},2^{-7},...,2^{-3}\}$, and as suggested in the paper \cite{RELM}, we use the geometric mean (GM) of Acc (1 - HL) and F1 as a performance metric ($\text{GM}= \sqrt{\text{Acc} \times \text{F1}}$). Therefore, we draw a 3-D network graph to represent the performance varying with the joint distribution of $\beta$ and $\gamma$. We choose two datasets, rcv1$\_$subset5 and medical, as examples to analyze the sensitivity of the parameters. 

Since two figures in Fig. \ref{Sens} have similar trends, we analyze Fig. \ref{Sens1} as a representative. As shown in Fig. \ref{Sens1}, as the value of $\beta$ decreases from the top, the performance continues to increase until 
$\beta = 0.55$ regardless of the value of $\gamma$. As $\beta$ decreases from $0.55$ to $0.5$, the performance drops sharply and remains at a lower value for $\beta \leq 0.5$. The reason for this is that a large $\beta$ means that the optimal label score is mostly determined by the observed labels, so it is still useful to further exploit the neighbourhood information by decreasing $\beta$, while too small $\beta$ causes overfitting of the label reconstruction, leading to the significant performance degradation. In addition, when $\beta > 0.5$, we can clearly observe that the performance curves with respect to $\gamma$ peak at the middle point $\gamma = 2^{-6}$, indicating the optimal balancing weight between the label ranking term and the label scoring term. Therefore, $(\beta,\gamma) = (0.55,2^{-6})$ can be selected as the optimal values on the data set rcv1$\_$subset5. 

\subsection{Efficiency}
The time complexity of ELM-NCLD can be divided into four parts: 1) To initialize the online model, computing $\mathbf{P}_0,\mathbf{\Phi}_0, \mathbf{Z}_0$ totally costs $\mathcal{O}(N^2L+NL^2+L^3+L^2q)$, which can be further denoted as 
$\mathcal{O}(N^2L)$ by the fact of $L,q \ll N$;
2) In each round of updating the online model ELM-NCLD, the time complexity comes from the computations of the matrices $\mathbf{S}, \mathbf{R}, \mathbf{P}, \mathbf{\Phi}$ and $\mathbf{Z}$. To optimize the objective in Eq. \eqref{Re}, a QP (quadratic programming) problem has to be solved with respect to $\mathbf{S}$ with the time cost $\mathcal{O}(kN^2)$, where $k$ is the total number of iterations. Then the computation of $\mathbf{R}$ with $\mathbf{S}$ costs $\mathcal{O}(KN^2)$ because each row of $\mathbf{S}$ has at most $K$ non-zero elements. According to the update rule of $\mathbf{P}, \mathbf{\Phi}$ in Eq. \eqref{Update}, the time consumption of updating $\mathbf{P}$ and $\mathbf{\Phi}$ is denoted as $\mathcal{O}(N^2L+NL^2+L^3)$ and $\mathcal{O}(N^2L+NLq+L^2q)$ respectively, and the time of updating $\mathbf{Z}$ via Eq. \eqref{Z_Update} costs $\mathcal{O}(N^2L+NLq+L^2q)$. Hence, the total time cost for matrix computation is denoted as $\mathcal{O}(N^2(k+K+L))$ ($L,q \ll N$). 3) At the \emph{i}-th round, an independent ELM model is trained on $D_i$ to estimate the noisy posterior probability, whose time cost is $\mathcal{O}(NL^2+NLq+L^3+L^2q)$; 4) To detect the potential concept drift, obtaining $\hat{LCard}(D_{i})$ needs $N \times q$ times of computations. In summary, the time complexity for one update of ELM-NCLD is denoted as $\mathcal{O}(N^2(k+K+L))$.

\section{Limitations and Future Works}
Our work utilizes the given noisy rates to derive unbiased  objective of online model, which is usually unavailable in real-world applications. It is more reasonable and practical to estimate the noisy rate as the data chunks arrive sequentially, rather than with the direct access. Therefore, our further work is to design an online framework that aims to classify the data with noisy label distributions without knowing noisy rates in advance, but by estimating the noisy rates during the process of online classification.

\section{The proof}
In this section, we provide a comprehensive set of proofs for the propositions presented in the previous sections.

\subsection{Proof of Proposition 1}
In terms of Eq. \eqref{N2}, the optimization term w.r.t. each $o_{tj}$ can be formulated as: 

\begin{equation}
\begin{aligned}
{\operatorname{min}} &\frac{\beta}{2}\|o_{tj}-y_{tj}\|_2^2 + \frac{1-\beta}{2}
\| o_{tj} - \sum_{n} S_{t,n}o_{nj}   \|_2^2, n \in \mathcal{N}\left({t}\right)  \notag
\end{aligned} 
\end{equation}
of which the gradient w.r.t. each $o_{tj}$ is: $\nabla_{o_{tj}} = \beta (o_{tj}-y_{tj}) + (1- \beta) (o_{tj}-\sum_{n} S_{t,n}o_{nj}), n \in \mathcal{N}\left({t}\right), t \in [N]$. Since the convexity of the objective in Eq. \eqref{N2} indicates the Karush-Khun-Tucker condition \cite{Cvx}, for each optimal label scores $o_{tj}^*$ must meet $\nabla_{o_{tj}^*} = 0$, hence the proof is completed.
 
\subsection{Proof of Proposition 2}
\begin{equation}
\begin{aligned}
\label{RL_Derive}
&\mathbb{E}_{(\mathbf{x}_t,y_{tj},y_{tk}) \sim D_{G}} (f(y^{(t)}_{j,k}o^{(t)}_{j,k})) 
\\ = & 
\int_{\mathbf{x}_t} \sum_{y_{tj}}\sum_{y_{tk}}P_{D_G}(\mathbf{x}_t,y_{tj},y_{tk})f(y^{(t)}_{j,k}o^{(t)}_{j,k})\mathrm{d} \mathbf{x}_t
\\ = & 
\int_{\mathbf{x}_t} \sum_{y_{tj}}\sum_{y_{tk}}\frac{P_{D_G}(\mathbf{x}_t,y_{tj},y_{tk})}{P_{D}(\mathbf{x}_t,y_{tj},y_{tk})} P_{D}(\mathbf{x}_t,y_{tj},y_{tk})f(y^{(t)}_{j,k}o^{(t)}_{j,k})\mathrm{d} \mathbf{x}_t
\\ = & 
\mathbb{E}_{(\mathbf{x}_t,y_{tj},y_{tk}) \sim D} (\frac{P_{D_G}(y_{tj},y_{tk} \mid \mathbf{x}_t)}{P_{D}(y_{tj},y_{tk} \mid \mathbf{x}_t)}f(y^{(t)}_{j,k}o^{(t)}_{j,k}))
\\ = &
\mathbb{E}_{(\mathbf{x}_t,y_{tj},y_{tk}) \sim D} (\frac{P_{D_G}(y_{tj} \mid \mathbf{x}_t)}{P_{D}(y_{tj} \mid \mathbf{x}_t)} \cdot \frac{P_{D_G}(y_{tk} \mid \mathbf{x}_t)}{P_{D}(y_{tk} \mid \mathbf{x}_t)}f(y^{(t)}_{j,k}o^{(t)}_{j,k}))
\\ = &
\mathbb{E}_{(\mathbf{x}_t,y_{tj},y_{tk}) \sim D} (\omega_{t,j}\omega_{t,k}f(y^{(t)}_{j,k}o^{(t)}_{j,k})) \notag
\end{aligned}
\end{equation}

The third equality holds because $D_G$ and $D$ share the same feature distribution (i.e., $P_{D_G}(\mathbf{x}_t) = P_{D}(\mathbf{x}_t)$), the fourth equality holds based on the assumption that $y_{tj},y_{tk}$ are conditionally independent given $\mathbf{x}_t$.

\subsection{Proof of Proposition 3}
\begin{equation}
\begin{aligned}
\label{Card_Derive}
&\mathbb{E}_{(\mathbf{x}_t,\mathbf{y}_{t}) \sim D_{G}} (LCard(\mathbf{x}_t,\mathbf{y}_{t})) 
\\ = & 
\int_{\mathbf{x}_t} \sum_{\mathbf{y}_{t}}P_{D_G}(\mathbf{x}_t,\mathbf{y}_{t}) \sum_{j=1}^q \mathbb{I}_{\{ y_{tj} = 1 \}}\mathrm{d} \mathbf{x}_t
\\ = & 
\int_{\mathbf{x}_t} P_{D_G}(\mathbf{x}_t) \sum_{\mathbf{y}_{t}}   \sum_{j=1}^q\mathbb{I}_{\{ y_{tj} = 1 \}} P_{D_G}(y_{tj} \mid \mathbf{x}_t)  P_{D_G}(\overline{\mathbf{y}}_{tj} \mid \mathbf{x}_t)   \mathrm{d} \mathbf{x}_t
\\ = & 
\int_{\mathbf{x}_t} P_{D_G}(\mathbf{x}_t) \sum_{j=1}^q   \sum_{{y}_{tj} } \mathbb{I}_{\{ y_{tj} = 1 \}} P_{D_G}(y_{tj} \mid \mathbf{x}_t)  \mathrm{d} \mathbf{x}_t
\\ = & 
\int_{\mathbf{x}_t} P_{D}(\mathbf{x}_t) \sum_{j=1}^q   \sum_{{y}_{tj} } 
 \mathbb{I}_{\{ y_{tj} = 1 \}} \frac{P_{D_G}(y_{tj} \mid \mathbf{x}_t)}{P_{D}(y_{tj} \mid \mathbf{x}_t)}
P_{D}(y_{tj} \mid \mathbf{x}_t)  \mathrm{d} \mathbf{x}_t
\\ = & 
\sum_{j=1}^q  \int_{\mathbf{x}_t} \sum_{{y}_{tj} }  P_{D}(\mathbf{x}_t,y_{tj})  
 \mathbb{I}_{\{ y_{tj} = 1 \}} \frac{P_{D_G}(y_{tj} \mid \mathbf{x}_t)}{P_{D}(y_{tj} \mid \mathbf{x}_t)}
 \mathrm{d} \mathbf{x}_t
\\ = & 
\sum_{j=1}^q \mathbb{E}_{(\mathbf{x}_t,y_{tj}) \sim D} \left(\mathbb{I}_{\{ y_{tj} = 1 \}} \omega_{t,j}\right) \notag
\end{aligned}
\end{equation}

\subsection{Proof of Proposition 4}
The Hoeffding inequality provides an upper bound on the probability that the sum of bounded independent random variables will deviate from its expected value by more than a certain amount, which states that:

Consider $X_1, X_2, \ldots, X_n$ as independent random variables, each taking values in a bounded interval $\left[a_i, b_i\right]$ for $i=1,2, \ldots, n$. Let $M_n=\sum_{i=1}^n X_i/n$ be their mean, and $\mu=\mathbb{E}\left[M_n\right]$ be the expected value of the mean. Then for any $\varepsilon>0$:
$$
P\left(\left|M_n-\mu\right| \geq \varepsilon\right) \leq 2 \exp \left(-\frac{2 n^2 \varepsilon^2}{\sum_{i=1}^n\left(b_i-a_i\right)^2}\right)
$$

Recall that the ground-truth cardinality w.r.t the \emph{t}-th instance in $D_i$ is estimated as $\hat{LCard}(\mathbf{x}^i_t) = 
 \sum_{j=1}^q \mathbb{I}_{\{ y_{tj} = 1 \}} \omega_{t,j}$. For the data chunk $D_i$ with $n = N$ observations, the mean cardinality is calculated as $M_n = \hat{LCard}(D_i) = 
\frac{1}{N}\sum_{t=1}^N \hat{LCard}(\mathbf{x}^i_t)$, since the exact upper bound and lower bound of cardinality w.r.t. each instance in $D_i$ are unavailable due to the label noise, we set $a_i = \min_{t}[\hat{LCard}(\mathbf{x}^i_t)], b_i = \max_{t}[\hat{LCard}(\mathbf{x}^i_t)]$ as approximation. By Additionally regarding the $\hat{LCard}(D_{i-1})$ as the expected cardinality of $D_i$, the proof is achieved by applying the the Hoeffding inequality.

\section*{Acknowledgment}
This work is supported in part by the Natural Science Foundation of China under grants (61976077, 62076085, 62106262, 62120106008), and the Fundamental Research Funds for the Central Universities of China under grant PA2022GDSK0038.

\bibliographystyle{IEEEtran}
\bibliography{myrefs}

\end{document}